\documentclass{article}

\usepackage{amssymb,amsfonts,amsmath,amsthm}
\usepackage{arxiv}

\usepackage[utf8]{inputenc} % allow utf-8 input
\usepackage[T1]{fontenc}    % use 8-bit T1 fonts
\usepackage{hyperref}       % hyperlinks
\usepackage{url}            % simple URL typesetting
\usepackage{booktabs}       % professional-quality tables
\usepackage{amsfonts}       % blackboard math symbols
\usepackage{nicefrac}       % compact symbols for 1/2, etc.
\usepackage{microtype}      % microtypography
\usepackage{lipsum}
\usepackage{graphicx}
\graphicspath{ {./images/} }

\title{Point Cloud Registration using Representative Overlapping Points}

\author{
 Lifa Zhu \\
  DeepGlint \\
  Beijing, China\\
  \texttt{lifazhu@deepglint.com} \\
   \And
 Dongrui Liu \\
  Shanghai Jiao Tong University\\
  Shanghai, China \\
  \texttt{drliu96@sjtu.edu.cn} \\
  \And
 Changwei Lin \\
  DeepGlint \\
  Beijing, China\\
  \texttt{changweilin@deepglint.com} \\
  \And
 Rui Yan \\
  Institute of Computing Technology, Chinese Academy of Sciences\\
  Beijing, China \\
  \texttt{yanrui20b@ict.ac.cn} \\
  \And
 Francisco Gómez-Fernández \\
  DeepGlint \\
  Beijing, China\\
  \texttt{francisco@deepglint.com} \\
  \And
 Ninghua Yang \\
  DeepGlint \\
  Beijing, China\\
  \texttt{ninghuayang@deepglint.com} \\
   \And
   Ziyong Feng \\
   DeepGlint \\
   Beijing, China\\
   \texttt{ziyongfeng@deepglint.com} \\
}

\begin{document}
\maketitle
\begin{abstract}
3D point cloud registration is a fundamental task in robotics and computer vision. Recently, many learning-based point cloud registration methods based on correspondences have emerged. However, these methods heavily rely on such correspondences and meet great challenges with partial overlap. In this paper, we propose ROPNet, a new deep learning model using Representative Overlapping Points with discriminative features for registration that transforms partial-to-partial registration into partial-to-complete registration. Specifically, we propose a context-guided module which uses an encoder to extract global features for predicting point overlap score. To better find representative overlapping points, we use the extracted global features for coarse alignment. Then, we introduce a Transformer to enrich point features and remove non-representative points based on point overlap score and feature matching. A similarity matrix is built in a partial-to-complete mode, and finally, weighted SVD is adopted to estimate a transformation matrix. Extensive experiments over ModelNet40 using noisy and partially overlapping point clouds show that the proposed method outperforms traditional and learning-based methods, achieving state-of-the-art performance. The source code is publicly available at \url{https://github.com/zhulf0804/ROPNet}.
\end{abstract}

\section{Introduction}
\label{sec:intro}

Rigid point cloud registration is a fundamental task in computer vision and robotics, aiming to find the rigid transformation that aligns a pair of point clouds. It has many important applications in scene reconstruction~\cite{blais1995registering, choi2015robust, merickel19883d}, positioning and localization~\cite{lu2019l3, wan2018robust}, object pose estimation~\cite{wong2017segicp} and so on.
Partially overlapping point cloud registration, i.e., partial-to-partial registration, is a common task in the real world. 
%FGF: rephrased
%Point clouds generated by depth cameras are always partially overlapping due to the camera shake. 
Robotic platforms equipped with modern depth cameras most of the time produce partially overlapping 3D point clouds due to the small camera motion between consecutive frames.
%FGF: rephrased
%In addition, the 3D reconstruction task needs to generate partially overlapping point clouds for registration.
In addition, in 3D reconstruction tasks, in order to produce a complete 3D scene, overlapping point clouds are captured from different  shooting angles.
%FGF: moved above
%Partially overlapping point cloud registration, i.e., partial-to-partial registration, is common task in the real world. 

Traditional methods, such as ICP~\cite{121791}, Go-ICP~\cite{7368945}, and FGR~\cite{zhou2016fast}, are widely used for rigid point cloud registration. However, they are sensitive to noise and partially overlapping point clouds. 
Recently, learning-based methods~\cite{aoki2019pointnetlk, sarode2019pcrnet, wang2019deep, wang2019prnet, li2019iterative, yew2020rpm, huang2020registration, yuan2020deepgmr, fu2021robust} have shown to be more robust, especially for partial-to-partial registration. 
%FGF: rephrased
%PRNet~\cite{wang2019prnet}, IDAM~\cite{li2019iterative} and RPMNet~\cite{yew2020rpm} solve partial-to-partial registration by generating sharp keypoints mapping matrix with Gumbel-Softmax~\cite{jang2016categorical}, a hybrid point elimination with mutual-supervised loss, and iterative Sinkhorn~\cite{sinkhorn1964relationship} layers to generate a doubly stochastic matrix, respectively. 
PRNet~\cite{wang2019prnet}, IDAM~\cite{li2019iterative} and RPMNet~\cite{yew2020rpm} solve partial-to-partial registration in different ways: by generating a sharp matching function based on keypoints with Gumbel-Softmax~\cite{jang2016categorical}; using a hybrid point elimination with a mutual-supervised loss, and applying iterative Sinkhorn~\cite{sinkhorn1964relationship} layers to generate a doubly stochastic matrix. 
%FGF: repharese
%None of them explicitly deal with non-overlapping points, which are critical for registration. 
However, none of them explicitly deal with \textit{non-overlapping points}, i.e. points without a matching pair between source and target point clouds, which are critical for registration. 
To the best of our knowledge, PREDATOR~\cite{huang2020registration} is the first work to explicitly deal with non-overlapping points.
%FGF: rephrase
%However, they unnecessary remove non-overlapping points from both source and target point clouds.
%Besides, when some overlapping points from the target point cloud are mistakenly removed, it can cause false correspondences which have adverse effect on registration.
% zhulifa rm: in this method
Despite its good performance, when some overlapping points are mistakenly removed from the target point cloud, false correspondences may be produced affecting the registration.

%FGF: the caption text is too long
\begin{figure}
\centering
\includegraphics[width=0.675\linewidth]{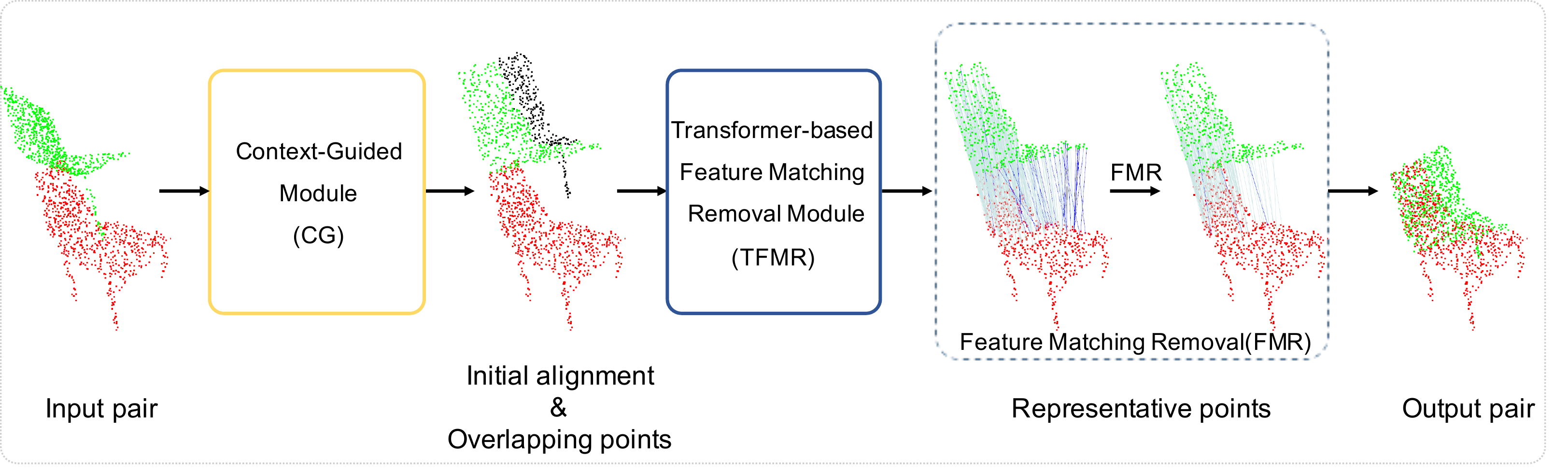}
\hfill
\includegraphics[width=0.32\linewidth]{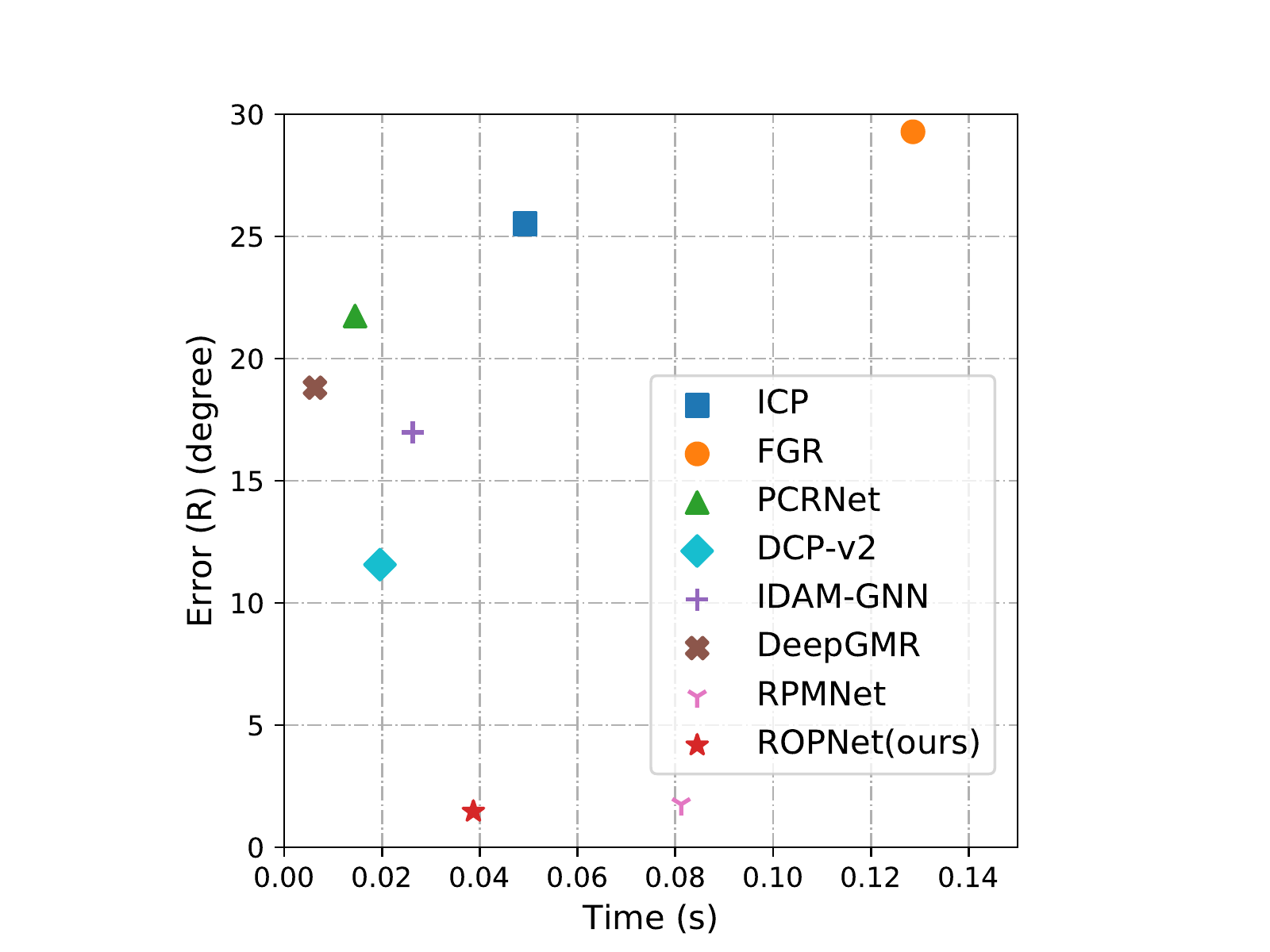}
\caption{Left: Overview of ROPNet registration pipeline. The CG module consumes the source (green) and target (red) point clouds, and outputs initial pose and overlapping points (non-overlapping points are in black). The TFMR module takes the output of CG module as input, and generates accurate correspondences.
%for the representative overlapping points. 
The FMR step removes false correspondences (blue lines) and keeps some positive correspondences (gray lines). Right: Error(R)-time comparison of different methods. Our ROPNet achieves the lowest $Error(R)$ with a moderate speed.}
\label{fig:ROPNet}
\end{figure}

Up to now, partial-to-partial registration is still not well resolved. Finding a few accurate correspondences is sufficient to solve the partial-to-partial registration problem. 
Although descriptive point features are essential to establish good point correspondences, they are always sensitive to rotation transformations. Even using rotation-invariant features, there may still be some false point correspondences due to the existence of noisy or non-overlapping points. 
Removing non-overlapping points in the source point cloud can effectively guarantee that each point in the source point cloud has a corresponding point in the target point cloud. 
We call this {\em partial-to-complete} registration where {\em complete} means the target point cloud is complete compared to the source point cloud.

Based on the above discussions, we propose \textit{ROPNet} for rigid partial-to-partial point cloud registration using representative overlapping points. 
First, we present a simple yet effective context-guided (CG) module to predict a point overlap score. The point overlap score is obtained through the information interaction between the point features and the global features from the pair. 
To better find representative overlapping points, we regress a rough initial transformation based on the global features. 
Second, we propose a Transformer-based feature matching removal (TFMR) module, which employs a point cloud Transformer for enriching point features and estimate a similarity matrix based on features to obtain representative overlapping points. 
Then, we remove non-representative points from the source point cloud and transform partial-to-partial registration to partial-to-complete registration. 
Finally, weighted SVD is employed to estimate the transformation matrix. Combining transformations from CG and TFMR modules, an accurate transformation matrix $T \in SE(3)$ are finally calculated.
The pipeline of our proposed ROPNet is shown in \autoref{fig:ROPNet}.

Extensive experiments showed that the proposed method achieves state-of-the-art performance on noisy and partially overlapping point clouds, outperforming traditional and learning-based methods.

%FGF: this paragraph is copy-pasted from the abstract. To rephrase or delete ?
%Extensive experiments over ModelNet40 using noisy and partially overlapping point clouds show that the proposed method outperforms traditional and learning-based methods, achieving state-of-the-art performance.

\section{Related Work}

\subsection{Traditional registration}

%TODO mention the least square problem
%FGF: rephrase
%The Iterative Closest Point (ICP) algorithm ~\cite{121791} alternates between estimating the correspondence by finding the closest point for every point and calculating the rigid transformation by solving the least square problem iteratively. 
The ICP algorithm~\cite{121791} iteratively alternates between two main steps until a desired stopping criteria is met: i) point correspondences computation between source and target point clouds, ii) rigid body transformation estimation by solving the least square problem. 
The variant point-to-plane ICP~\cite{chen1992object, rusinkiewicz2001efficient} uses a different objective function, which requires normal vectors for the target point cloud, that has been shown a better and faster convergence. 
%to represent the underlying surface
However, ICP and its variants are sensitive to initial alignment and noisy points, and also they easily fall into a local optimal.

Another family of registration algorithms, known as global registration, computes a rough transformation to align source and target point clouds~\cite{raguram2008comparative, 7368945, zhou2016fast}. 
The branch-and-bound methods, e.g. Go-ICP~\cite{7368945}, systematically search for the optimal solution in the pose space.
%FGF: rephrase
%RANSAC-based~\cite{raguram2008comparative} methods pick a fixed number of random points from the source point clouds and the corresponding points in the target point cloud in each iteration and evaluate this alignment. 
RANSAC-based~\cite{raguram2008comparative} methods randomly pick in each iteration a fixed number of points from the source and target point clouds to estimate an alignment, and finally output the best transformation.
However, the above methods are time-consuming, which is not acceptable in practical applications. 
%FGF: rephrase
%FGR~\cite{zhou2016fast} uses Fast Point Feature Histogram (FPFH)~\cite{rusu2009fast} descriptors to generate point correspondences and the correspondencesin the target point cloud are detected in the 33-dimensional FPFH feature space
FGR~\cite{zhou2016fast} extracts Fast Point Feature Histogram (FPFH)~\cite{rusu2009fast} descriptors 
to find good matching point pairs in a 33-dimensional feature space.
However, handcrafted features are sensitive to noise and FGR doesn't work well on partial-to-partial registration.

\subsection{Learning-based registration}

Global features based methods, such as PointNetLK~\cite{aoki2019pointnetlk} and PCRNet~\cite{sarode2019pcrnet}, both use PointNet~\cite{qi2017pointnet} to extract global features. PointNetLK uses a modified Lucas \& Kanade (LK) algorithm~\cite{lucas1981iterative} for aligning features to achieve registration. PCRNet replaces the LK algorithm with Multi-Layer Perceptron (MLP) to promote the robustness to the noisy points. They both capture global features for complete-to-complete registration, causing massive information loss and inaccurate transformations. 

%FGF: include subsection/paragraph "correspondence-based methods" ?
% zhulf: i try it
{\bf Correspondence-based} Deep Closest Point (DCP)~\cite{wang2019deep} learns point features, computes a distance matrix based on features, and generates virtual correspondences to solve the least square problem.
DeepGMR~\cite{yuan2020deepgmr} solves the point correspondence step by matching points using a Gaussian Mixture Model whose parameters are estimated by a neural network. 
PRNet~\cite{wang2019prnet} performs partial-to-partial registration by detecting keypoints and establishing keypoint-to-keypoint correspondences with Gumbel-Softmax~\cite{jang2016categorical}. 
RPMNet~\cite{yew2020rpm} solves partial-to-partial registration by applying iterative Sinkhorn~\cite{sinkhorn1964relationship} layers to generate a doubly stochastic matrix~\cite{sinkhorn1964relationship}. 
IDAM~\cite{li2019iterative} incorporates both features and Euclidean information for correspondence matrix and uses a two-stage point elimination for partial-to-partial registration. 
%FGF: rephrase
%The above correspondence-based methods do not consider the partial-to-partial problem or solve the problem indirectly.
The above correspondence-based methods solves the partial-to-partial registration problem indirectly or do not consider it at all.

%FGF: include RANSAC-based methods section or paragraph ?
{\bf RANSAC-based} FCGF \cite{choy2019fully} uses a UNet \cite{ronneberger2015u} architecture with skip connections and residual blocks to extract fully-convolutional features. 
D3Feat~\cite{bai2020d3feat} applies KPConv~\cite{thomas2019kpconv} as backbone to predict features and point saliency. PREDATOR~\cite{huang2020registration} proposes an overlap-attention block for point feature extraction, overlap score and saliency prediction. 
However, the above RANSAC-based methods tend to be slow. In addition, mistakenly removal of non-overlapping points from the target point cloud may produce false matches.

\section{Methods}

%FGF: rephrase
%Given two input point clouds $X \in \mathbb R^{N \times 3}$, $Y \in \mathbb R^{M \times 3}$, rigid point cloud registration aims to find the rigid body transformation $R$, $t$ that aligns the two point clouds.
The partial-to-partial rigid point cloud registration problem can be stated as follows.
Given two input point clouds $X \in \mathbb R^{N \times 3}$ and $Y \in \mathbb R^{M \times 3}$, our goal is to find a rigid body transformation $R$, $t$ that better aligns $X$ to $Y$ in the least-squares sense using a learned subset of overlapping point correspondences.

%FGF: moved to the end of section 3.3
% Based on the ground truth transformation $\hat R$, $\hat t$, we define the overlapping point set for $X$ as:
% \begin{equation}
% \label{eq:1}
% \begin{aligned}
% \Phi_X = \{X_i | NN(\hat R \cdot X_i + \hat t, Y) < d, \forall i\}, 
% \end{aligned}
% \end{equation}
% where $NN(x, Y)$ denotes finding the nearest neighbour in $Y$ for the point $x$, and $d$ is a predefined threshold. 

\subsection{Context-guided module for initialization and overlap}

\begin{figure}[t]
\centering
\includegraphics[width=0.9\textwidth]{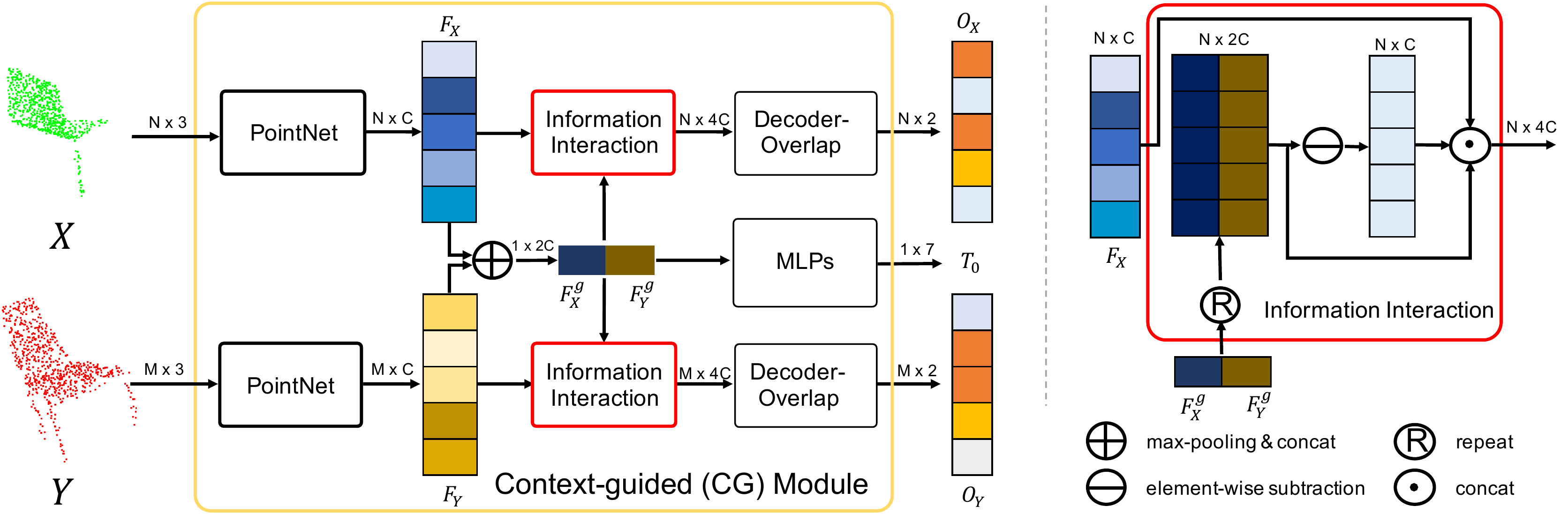} \\
\caption{Left: Architecture of the CG module. CG module consumes source $X$ (in green) and target $Y$ (in red) data, and outputs overlap score ($O_X$, $O_Y$) and initial transformation matrix $T_{0}$. Right: Details of information interaction. It takes point features and global features as input and outputs fused point features based on the pair.}
\label{fig:ROP_CG}
\end{figure}

%FGF: rephrase
%A rough initial pose is conducive to point features learning due to small rotation transformation. 
We have observed that a rough initial registration is always favourable to point feature learning due to small rotation transformation. 
In SpinNet~\cite{ao2021spinnet}, it is shown that rotation-invariant features are hard to learn.
Also, overlapping points can greatly improve the correspondence accuracy that are critical for registration \cite{huang2020registration}. 
Therefore, we propose a context-guided (CG) module for initialization and point overlapping score prediction. \autoref{fig:ROP_CG} shows the CG architecture.

{\bf Initialization}
%FGF: rephrase
%For simplicity and rough initial alignment estimation, we use a modified PointNet network as the encoder of our CG module. 
For simplicity, we use a modified PointNet network as the encoder of our CG module. 
%Still, it may also be implemented by the other formulations of point cloud convolution. 
Inspired by PCRNet~\cite{sarode2019pcrnet}, we use the encoder module to extract global features and regress a 7-dimensional vector representing a rotation and translation transformation. Then, the initial alignment can be formulated as:
\begin{equation}
v = h_\phi(\text{cat}(\max(f_\theta(X)), \max(f_\theta(Y)))),
\end{equation}
where $f_\theta(\cdot)$ denotes the shared encoder module which 
%consumes the source $X \in \mathbb R^{N \times 3}$ and target point cloud $Y \in \mathbb R^{M \times 3}$, and 
learns a high-dimensional feature $F_X \in \mathbb R^{N \times C}$ and $F_Y \in \mathbb R^{M \times C}$ for the source $X$ and target $Y$ input point clouds.
The symbol $\max(\cdot)$ denotes max-pooling 
%which outputs global features $F_X^g \in \mathbb R^{C}$ and $F_Y^g \in \mathbb R^{C}$ for input pair,
and $\text{cat}$ represents concatenation. The transformation decoder is denoted as $h_\phi(\cdot)$ which is a simple Multiple-Layer Perceptron (MLP). 
%FGF: remove following paragraph (it's redundant)
%Finally, the 7-dimensional vector $v$ is obtained where the 
The first fourth values of $v$ represent the quaternion rotation vector $q_0 \in \mathbb R^4$ that are used to calculate the initial transformation matrix $R_{0}$, and the last three values represent the translation $t_{0} \in \mathbb R^3$. 
%The initial transformation matrix $R_{0}$ is calculated from quaternion $q_0$. 
%The initial alignment $X'$ for $X$ can be obtained by $X \cdot R_{0}^T + t_{0}$.
The transformed source point cloud $X'$ can be obtained by $X' = X \cdot R_{0}^T + t_{0}^T$, abusing notation for the summation of $t_0$ which is done row by row.

{\bf Overlap} Information from both source and target point cloud is essential for overlapping points prediction. 
%FGF: removed
%Intuitively, humans glance at a point cloud pair and instantly know the overlapping points between them. 
Thus, we design a simple information interaction module for overlapping points prediction based on global features. These features are obtained from the initialization step, saving computation. Here, we formulate the overlap prediction problem as a binary classification task. The overlap score can be calculated as:
\begin{equation}
\begin{aligned}
O_X &= g_\psi(\text{cat}(F_X, r(F_X^g), r(F_Y^g), r(F_X^g - F_Y^g))), \\
O_Y &= g_\psi(\text{cat}(F_Y, r(F_Y^g), r(F_X^g), r(F_Y^g - F_X^g))),
\end{aligned}
\end{equation}
where $r(\cdot)$ denotes expanding dimensions and repeating features, $g_\psi$ is the overlap decoder (which is denoted in a PointNet-style) followed by a softmax function, and $O_X$ and $O_Y$ denote point overlap score of the source $X$ and target $Y$ point clouds. 
The parameters from $g_\psi$ are shared among input point clouds.

\subsection{Transformer-based feature matching removal}
\label{subsec:tfmr}

\begin{figure}[t]
\centering
\includegraphics[width=0.9\textwidth]{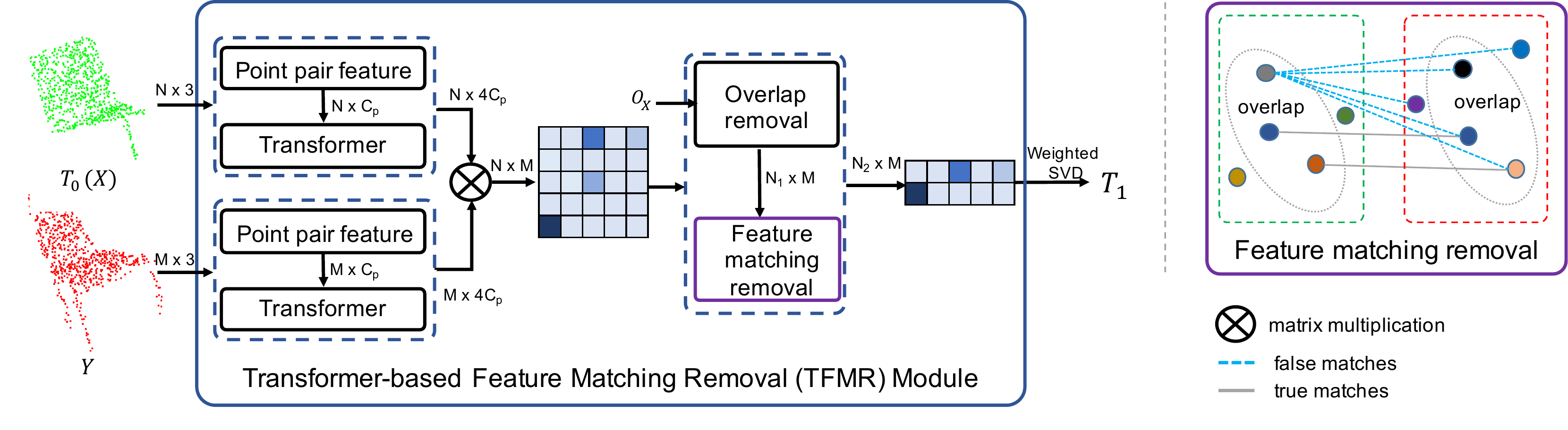} \\
\caption{Left: Overview of Transformer-based feature matching removal (TFMR) module. TFMR module takes the transformed source $X'$ and target $Y$ as input, and outputs representative points and their correspondences. $T_0$ and $O_X$ are the output from CG module that denote initial alignment and overlap score for $X$.
%FGF: removed
%$C_p$ denotes the feature dimension of the input for the Transformer module
Right: Details of feature matching removal (FMR). It takes correspondences for overlapping source points and outputs accurate correspondences (gray lines).}
\label{fig:ROP_TFMR}
\end{figure}

We refine the registration based on the point correspondences from transformed source $X'$ and target $Y$. The proposed Transformer-based feature matching removal (TFMR) module removes non-representative points from $X'$. The architecture of the TFMR module is depicted in \autoref{fig:ROP_TFMR}. 
%FGF: here

{\bf Point pair feature} Before Transformer, we implement point feature augmentation following RPMNet~\cite{yew2020rpm}. For each point $x \in \mathbb X'$, we obtain the k-nearest neighbor points $M_k \in \mathbb R^{k \times 3}$ within a predefined radius. 
We denote the augmented feature as $F^a_{m_i} \in \mathbb R^{10}$ for $m_i \in M_k$ which concatenates spatial coordinates and Point Pair Feature~\cite{drost2010model} $PPF_i \in \mathbb R^4$.
Then we get the feature $F_x^p \in \mathbb R^{C_p}$ for the point $x$ as: 
\begin{equation}
\begin{aligned}
F^a_{m_i} &= [m_i, m_i - x, PPF_i], \quad F^p_x = \max_{1 \leq i \leq k} (\mu_{\theta}(F^a_{m_i})),
\end{aligned}
\end{equation}
where $\mu_{\theta}$ consists of several 1-d convolution layers and $\max$ denotes max-pooling.

{\bf Transformer} Point features require to consider global information. Transformer~\cite{guo2020pct, zhao2020point, engel2020point} integrates point features based on all input points, showing advantages over local features~\cite{qi2017pointnet++, wang2019dynamic, thomas2019kpconv, liu2019relation, li2018pointcnn, zhao2019pointweb}. 
Given an augmented feature $F^p_{X'} \in \mathbb R^{N \times C_p}$ (abbreviated as $F_{X'}$ for convenience) for the transformed source point cloud $X'$, we generate point features based on self-attention as follows:
\begin{equation}
\begin{aligned}
& Q_i = F^i_{X'} \cdot W_i^Q, \quad K_i = F^i_{X'} \cdot W_i^K, \quad V_i = F^i_{X'} \cdot W_i^V, \\
& s_i = \text{norm}(\text{softmax}(Q_i \cdot K_i^T)), \quad F_{X'}^{i+1} = s_i \cdot V_i, \\
\end{aligned}
\end{equation}
where $\cdot$ denotes matrix multiplication, $i$ is the index of self-attention and $F_{X'}^0 = F_{X'}$, $W_i^Q \in \mathbb R^{C_p \times C_0}$, $W_i^K \in \mathbb R^{C_p \times C_0}$ and $W_i^V \in \mathbb R^{C_p \times C_p}$ are learned weights  and $C_0$ is a hyperparameter used to control the amount of calculation.
%FGF: $C_0$ is a hyper-parameter ?
%zhulf: yes
The symbol $\text{softmax}(\cdot)$ denotes row softmax and $\text{norm}(\cdot)$ means column normalization. Following self-attention, the Feature Forward Neural Network (FFNN) is a residual network. Then, Transformer is composed of several self-attention and FFNN groups. The output features for $X'$ and $Y$ are represented as $F^t_{X'} \in \mathbb R^{N \times 4C_p}$ and $F^t_{Y} \in \mathbb R^{M \times 4C_p}$.

{\bf Feature matching removal} We first compute the similarity matrix $H = F_{X'}^t \cdot {F_Y^t}^T$. However, there may exist points in $X'$ without a corresponding point in $Y$, such that the correspondence matrix satisfies $\sum_j^M{H_{ij}} \leq 1, \forall  i$. 
We can remove non-overlapping points in $X'$ which guarantees the remaining points $X'_{o1}$ to satisfy $\sum_j^M{H_{o1, ij}} = 1, \forall i$, so it is sufficient for solving partial-to-partial registration. 
In this way, we obtain representative overlapping points in two steps. First, we select top-$N_1$ points as overlapping points $X'_{o1} \in \mathbb R^{N_1 \times 3}$ based on the overlap score $O_X$ predicted in the CG module. Second, we perform feature matching to further remove points whose features are not descriptive to obtain the final representative points:
\begin{equation}
\begin{aligned}
X'_{o2} = X'_{o1}[\underset{ \text{top-}prob}{\arg \max}(\max_j(F_{X'_{o1}}^t \cdot {F_Y^t}^T)_{ij})], 
\end{aligned}
\label{eq:5}
\end{equation}
where top-$prob$ means obtaining an amount of indices with top similarity scores in the probability proportion of $prob$. We then obtain the similarity matrix $H_{o2} = F_{X'_{o2}}^{t} \cdot {F_Y^t}^T$ for representative overlapping points which transforms partial-to-partial registration into partial-to-complete registration. For each point $x_{o2, i} \in X'_{o2}$, the corresponding point $y_i$ is computed as following:
\begin{equation}
\begin{aligned}
J = \underset{\text{top}-k}{\arg \max} H_{o2, i}, \quad
w_{ij} =
\begin{cases}
H_{o2, ij} & j \in J \\
0 & j \notin J
\end{cases}, \quad y_{i} = \frac{w_{i, :}}{\sum_j w_{i, j}} \cdot Y,
\end{aligned}
\end{equation}
where $J$ denotes the indices of $Y$ which have $\text{top-}k$ maximum similarity scores for $x_{o2, i}$. 
To generate the set of correspondences, each point pair is defined as $(x_{o2, i}, y_i)$ with weight $O_{X'_{o2}, i}$.
Then, weighted SVD is used for estimating an accurate transformation $R_{1} \in \mathbb {SO}(3)$ and $t_{1} \in \mathbb R^{3}$. 
The final transformation is obtained by setting $R = R_{1} \cdot R_{0}$, and $t = R_{1} \cdot t_{0} + t_{1}$. 

\subsection{Loss function}

We adopt the widely used loss function which calculates the distance between predicted and ground truth transformed source data. 
%FGF:rephrase
%As we regard overlap score prediction as a binary classification task, we supervise it with cross-entropy loss. Besides, to make training more efficient, we adopt the same supervised signal as $R, t$ for initialization $R_0, t_0$ as an auxiliary loss.
Since we consider the prediction of the overlap score as a binary classification task, we supervise it with a cross-entropy loss.
In addition, to make training more efficient, we adopt the same loss as $R, t$ for initial $R_0, t_0$ as an auxiliary loss:
\begin{equation}
\begin{aligned}
& L_{fin} = ||X \cdot R^T  - X \cdot {\hat R}^T||_1 +  ||t - \hat t||_1, \quad L_{init} = ||X \cdot R^T_{0} - X \cdot {\hat R}^T||_1 +  ||t_{0} - \hat t||_1, \\
& L_{ol} = \frac{1}{2N} \sum_i \sum_j (\hat O_{X})_{ij} \cdot \log (O_X)_{ij} + \frac{1}{2M} \sum_i \sum_j (\hat O_{Y})_{ij} \cdot \log (O_Y)_{ij},\\
\end{aligned}
\end{equation}
where $\hat R$ and $\hat t$ denotes the ground truth rotation matrix and translation vector, and $\hat O_{X}$ and $\hat O_{Y}$ represent the ground truth overlap score using the mapping $\Phi_X$ obtained by \autoref{eq:1}, 
\begin{equation}
\label{eq:1}
\begin{aligned}
\Phi_X = \{X_i | NN(\hat R \cdot X_i + \hat t, Y) < d, \forall i\}, 
\end{aligned}
\end{equation}
where $NN(x, Y)$ denotes finding the nearest neighbour in $Y$ for the point $x$. Then, we set the score to 1 for correspondences below a predefined threshold $d$ and 0 otherwise.

\subsection{Implementation}

ROPNet is implemented in PyTorch~\cite{paszke2017automatic, paszke2019pytorch} and all experiments were run on a Tesla V100 GPU with an Intel 6133 CPU at 2.50GHz CPU.
For the CG module, the number of filters in the  encoder are $[64, 64, 64, 128, 512]$. The number of filters in transformation and overlap decoder are $[512, 512, 256, 7]$ and $[512, 512, 256, 2]$, respectively. For the TFMR module, we have four self-attention and FFNN layers where each output has 192 dimensions.
We adopt Group Normalization (GN)~\cite{wu2018group} in the TFMR module. For the ModelNet40 dataset, we train for 600 epochs using Adam~\cite{kingma2014adam} with initial learning rate of 0.0001. The learning rate changes using a cosine annealing schedule~\cite{loshchilov2016sgdr}. We train ROPNet in a non-iterative manner. However, we run 2 iterations for the TFMR module during test.

\section{Experiments}

\subsection{Dataset}

{\bf ModelNet40}~\cite{wu20153d} is widely used for point cloud registration as conducted in recent learning-based methods~\cite{wang2019deep, sarode2019pcrnet, wang2019prnet, yew2020rpm, wei2020end, xu2021omnet}. There are 12,311 CAD models from 40 categories, spliting into 9,843 for training and 2,468 for testing. 

It is noted that there are some symmetrical objects in ModelNet40, including bottle, bowl, cone, cup, flower pot, lamp, tent, and vase~\cite{yew2020rpm, xu2021omnet}. It's unfair to use the following metrics to evaluate these symmetrical categories. To tackle this problem, we evaluate on total objects (TO) and the asymmetric objects (AO) by removing symmetrical objects in the following experiments, respectively.

\subsection{Evaluation metrics}

We evaluate the registration in terms of the isotropic rotation and translation error $Error(R) = \text{arccos}\frac{tr(\hat R^{-1}R) - 1}{2}, Error(t) = ||\hat R^{-1} t - \hat t||_1$ proposed in RPMNet\cite{yew2020rpm}, where $R, t$ and $\hat R, \hat t$ represent the predicted and the ground truth transformation respectively, $tr(\cdot)$ means the trace of matrix. Moreover, we evaluate an-isotropic rotation and translation error $MAE(R), MAE(t)$ used in DCP~\cite{wang2019deep} by calculating mean absolute error of Euler angle and translation vector. Both $Error(R)$ and $MAE(R)$ represent rotation error in degrees.

\subsection{Comparison with other methods}

We compare ROPNet against ICP and FGR~\cite{121791, zhou2016fast} and learning-based methods which include PCRNet, DCP-v2, IDAM-GNN, DeepGMR, and RPMNet~\cite{yew2020rpm, wang2019deep, sarode2019pcrnet, li2019iterative, yuan2020deepgmr}. We generate source and target point cloud $X$ and $Y$ in the style of twice sample by sampling 1024 points independently following RPMNet~\cite{yew2020rpm}. We randomly generate three Euler angles within $[0^{\circ}, 45^{\circ}]$ and translations within $[-0.5, 0.5]$ on each axis. For generating partial point cloud, we adopt the mode conducted in RPMNet~\cite{yew2020rpm} which projects all points to a random direction, and 30\% points are removed for partial-to-partial registration. The following experiments are all performed in the partial-to-partial setting to evaluate on both AO and TO data of ModelNet40.
\begin{table}
\begin{center}
\scalebox{1.0}{
\begin{tabular}{|l|c|c|c|c|c|c|c|c|}
\hline
 & \multicolumn{4}{|c|}{AO} &  \multicolumn{4}{|c|}{TO} \\
\hline
Methods & $Error(R)$ & $Error(t)$ & $MAE(R)$ & $MAE(t)$ & $Error(R)$ & $Error(t)$ & $MAE(R)$ & $MAE(t)$ \\
\hline
ICP & 23.4101 & 0.2493 & 11.6681 & 0.1147 & 24.6413 & 0.2525 & 12.1470 & 0.1171 \\
FGR & 14.3599 & 0.0927 & 8.4620 & 0.0440 & 14.0114 & 0.0978 & 8.2744 & 0.0462 \\
\hline
PCRNet & 24.7053 & 0.1997 & 11.3387 & 0.0966 & 24.3003 & 0.2000 & 11.2207 & 0.0969 \\
DCP-v2 & 10.3151 & 0.1319 & 5.2299 & 0.0640 & 11.1723 & 0.1356 & 5.6421 & 0.0657 \\
IDAM-GNN & 13.6264 & 0.1886 & 7.1060 & 0.0862 & 14.2891 & 0.1909 & 7.4966 & 0.0877 \\
DeepGMR & 13.3886 & 0.1542 & 6.6794 & 0.0751 & 14.3612 & 0.1589 & 7.0914 & 0.0775 \\
RPMNet & 0.9335 & 0.0113 & 0.4860 & 0.0054 & 1.4239 & 0.0139 & 0.7304 & 0.0065 \\
ROPNet & {\bf 0.7820} & {\bf 0.0086} & {\bf 0.4146} & {\bf 0.0040} & {\bf 1.1567} & {\bf 0.0108} & {\bf 0.5946} & {\bf 0.0051} \\
\hline
\end{tabular}}
\end{center}
\caption{Results on ModelNet40 unseen shapes.}
\label{table:1}
\end{table}
%------------------------------------------------------------------------- 

{\bf Unseen shapes} We use 40 categories for training, then test both TO and AO data on the test set. The results in \autoref{table:1} indicate that our ROPNet outperforms the other methods with $Error(R)$ 0.7820, $Error(t)$ 0.0086 on AO, and $Error(R)$ 1.1567, $Error(t)$ 0.0108 on TO. It also shows that RPMNet outperforms DVP-v2, IDAM-GNN, DeepGMR, and ICP. Moreover, our ROPNet model always remains superior being still more than twice as fast as RPMNet illustrated in \autoref{fig:ROPNet}.

\begin{table}
\begin{center}
\scalebox{1.0}{
\begin{tabular}{|l|c|c|c|c|c|c|c|c|}
\hline
 & \multicolumn{4}{|c|}{AO} &  \multicolumn{4}{|c|}{TO} \\
\hline
Methods & $Error(R)$ & $Error(t)$ & $MAE(R)$ & $MAE(t)$ & $Error(R)$ & $Error(t)$ & $MAE(R)$ & $MAE(t)$ \\
\hline
ICP & 23.7983 & 0.2512 & 12.0064 & 0.1180 & 25.4879 & 0.2504 & 12.8976 & 0.1183 \\
FGR & 15.6677 & 0.0934 & 9.6716 & 0.0443 & 15.3317 & 0.1036 & 9.4110 & 0.0491 \\
\hline
PCRNet & 21.0916 & 0.1900 & 10.5662 & 0.0929 & 21.5943 & 0.1915 & 10.7881 & 0.0933 \\
DCP-v2 & 10.4008 & 0.1490 & 5.4455 & 0.0723 & 11.1543 & 0.1564 & 5.7080 & 0.0758 \\
IDAM-GNN & 15.9915 & 0.2109 & 8.4099 & 0.0991 & 17.3614 & 0.2153 & 8.9823 & 0.1022 \\
DeepGMR & 17.4120 & 0.1939 & 8.7646 & 0.0951 & 18.3608 & 0.1998 & 9.1425 & 0.0977 \\
RPMNet & 1.1273 & 0.0143 & 0.5987 & 0.0069 & 1.6781 & 0.0169 & 0.8749 & 0.0079 \\
ROPNet & {\bf 0.9285} & {\bf 0.0114} & {\bf 0.5029} & {\bf 0.0054} & {\bf 1.1637} & {\bf 0.0116} & {\bf 0.6190} & {\bf 0.0055} \\
\hline
\end{tabular}}
\end{center}
\caption{Results on ModelNet40 unseen categories.}
\label{table:2}
\end{table}
%------------------------------------------------------------------------- 

{\bf Unseen categories} We use the first 20 categories for training and the rest 20 categories for testing to validate the model generalization ability.  As shown in \autoref{table:2}, ICP and FGR obtain similar performances. IDAM-GNN and DeepGMR obtain higher error showing slightly weaker generalization ability. DCP-v2, RPMNet, ROPNet and PCRNet obtain similar or lower error showing great generalization ability. Overall, ROPNet outperforms all the compared methods in both AO and TO data.

\begin{table}
\begin{center}
\scalebox{1.0}{
\begin{tabular}{|l|c|c|c|c|c|c|c|c|}
\hline
 & \multicolumn{4}{|c|}{AO} &  \multicolumn{4}{|c|}{TO} \\
\hline
Methods & $Error(R)$ & $Error(t)$ & $MAE(R)$ & $MAE(t)$ & $Error(R)$ & $Error(t)$ & $MAE(R)$ & $MAE(t)$ \\
\hline
ICP & 24.3766 & 0.2542 & 12.3258 & 0.1196 & 25.5219 & 0.2515 & 12.9118 & 0.1190 \\
FGR & 28.6681 & 0.1742 & 17.7317 & 0.0827 & 29.2813 & 0.1828 & 18.1986 & 0.1887 \\
\hline
PCRNet & 21.4136 & 0.1924 & 10.7220 & 0.0940 & 21.7476 & 0.1933 & 10.8870 & 0.0942 \\
DCP-v2 & 10.8202 & 0.1508 & 5.6436 & 0.0734 & 11.5672 & 0.1578 & 5.9095 & 0.0765 \\
IDAM-GNN & 16.6100 & 0.2206 & 8.7257 & 0.1033 & 16.9805 & 0.2125 & 8.9366 & 0.1003 \\
DeepGMR & 17.8840 & 0.1972 & 8.9990 & 0.0967 & 18.8075 & 0.2027 & 9.3672 & 0.0992 \\
RPMNet & 1.2343 & 0.0157 & 0.6544 & 0.0075 & 1.7569 & 0.0177 & 0.9177 & 0.0084 \\
ROPNet & {\bf 1.1566} & {\bf 0.0137} & {\bf 0.6215} & {\bf 0.0066} & {\bf 1.4656} & {\bf 0.0145} & {\bf 0.7799} & {\bf 0.0070} \\
\hline
\end{tabular}}
\end{center}
\caption{Results on ModelNet40 unseen categories with Gaussian noise.}
\label{table:3}
\end{table}
%------------------------------------------------------------------------- 

\begin{figure}
\scalebox{0.8}{
\begin{tabular}{cccccccccc}
Inputs & ICP & FGR & PCRNet & DCP & IDAM & DeepGMR & RPMNet &  ROPNet & G.T. \\
\includegraphics[width=0.1\textwidth]{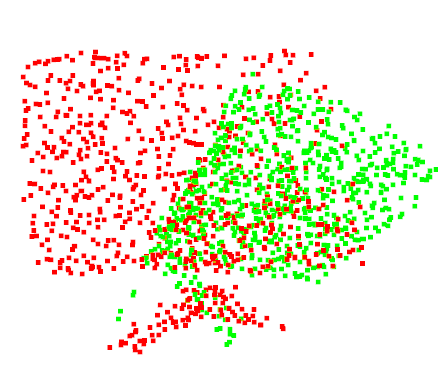} & \includegraphics[width=0.1\textwidth]{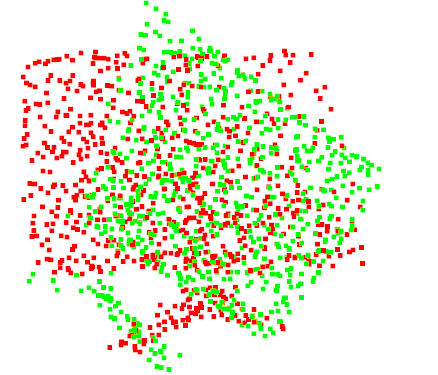} & \includegraphics[width=0.1\textwidth]{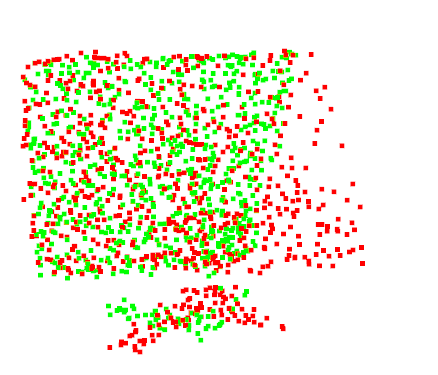} & \includegraphics[width=0.1\textwidth]{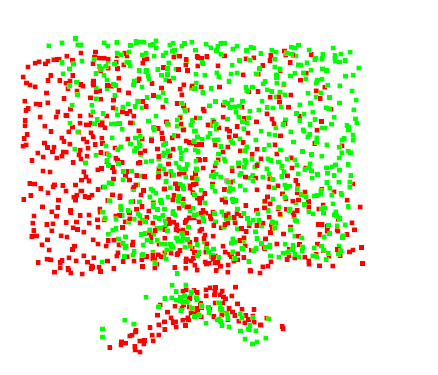} & \includegraphics[width=0.1\textwidth]{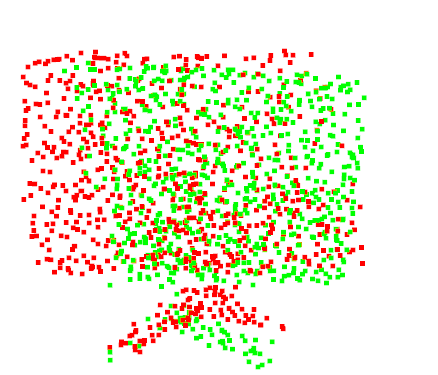} & \includegraphics[width=0.1\textwidth]{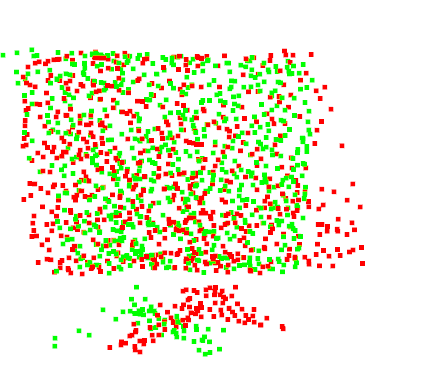} & \includegraphics[width=0.1\textwidth]{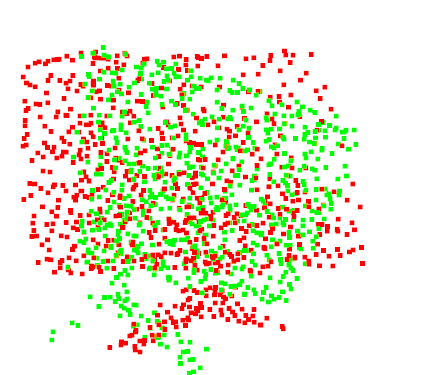} & \includegraphics[width=0.1\textwidth]{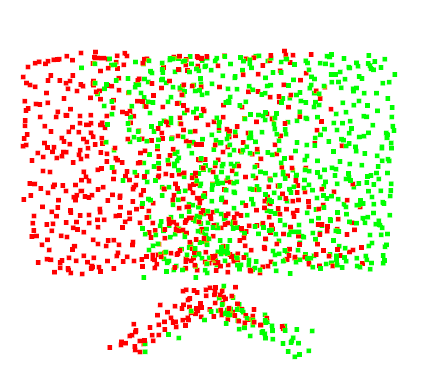} & \includegraphics[width=0.1\textwidth]{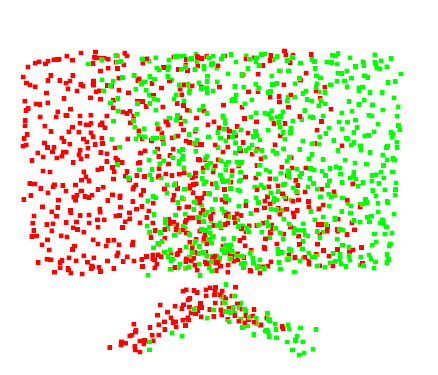} & \includegraphics[width=0.1\textwidth]{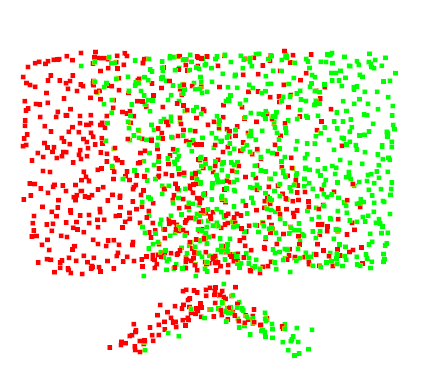} \\
\includegraphics[width=0.1\textwidth]{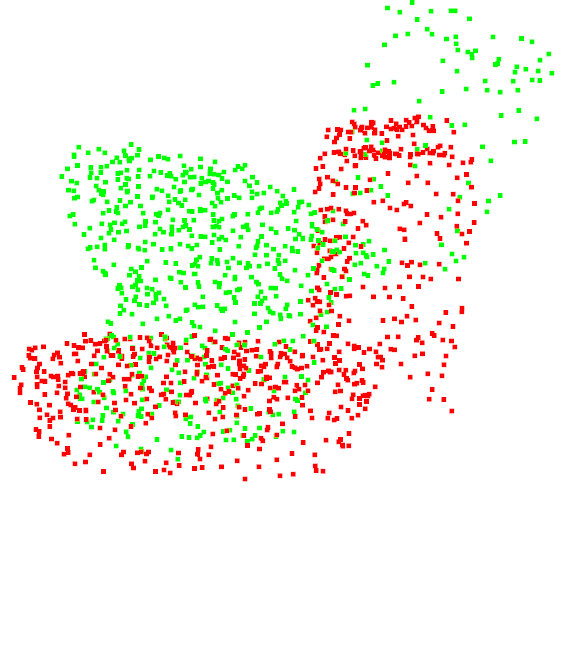} & \includegraphics[width=0.1\textwidth]{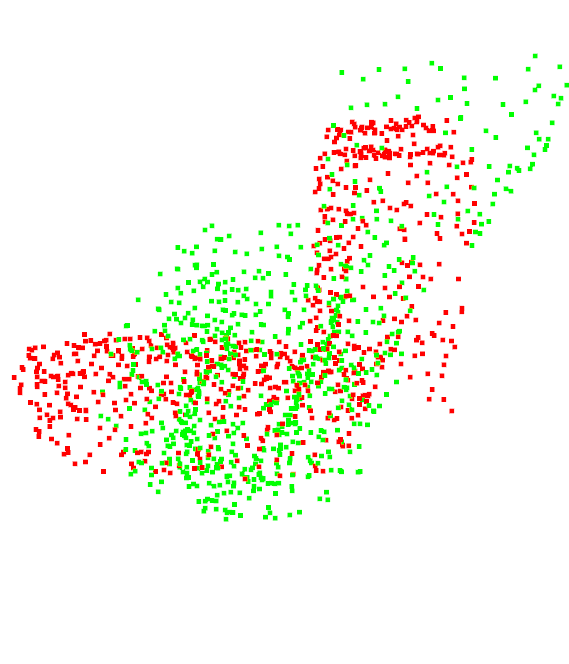} & \includegraphics[width=0.1\textwidth]{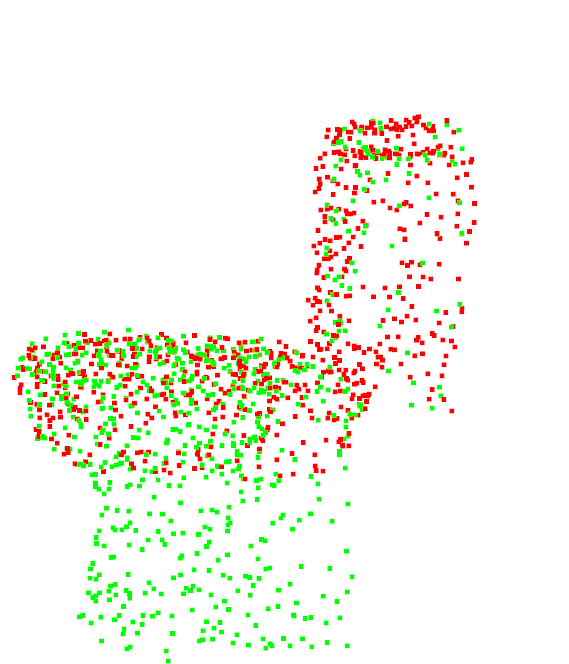} & \includegraphics[width=0.1\textwidth]{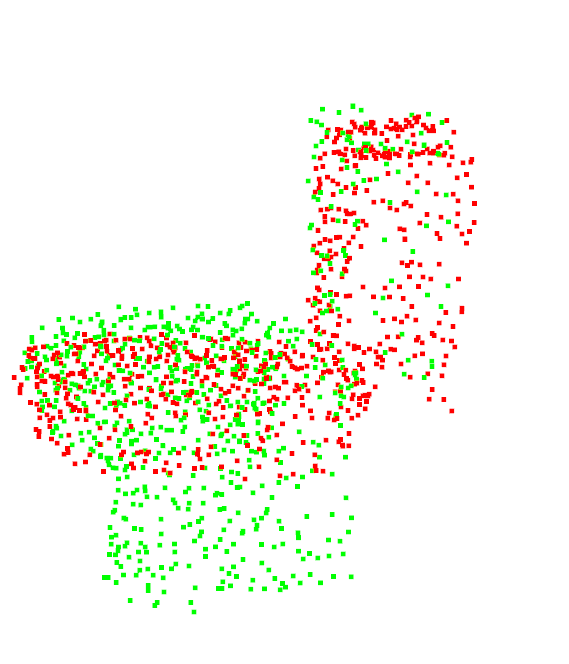} & \includegraphics[width=0.1\textwidth]{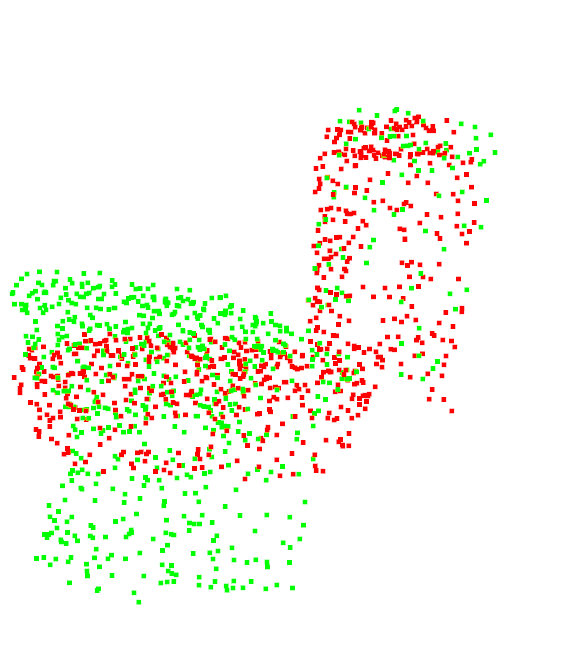} & \includegraphics[width=0.1\textwidth]{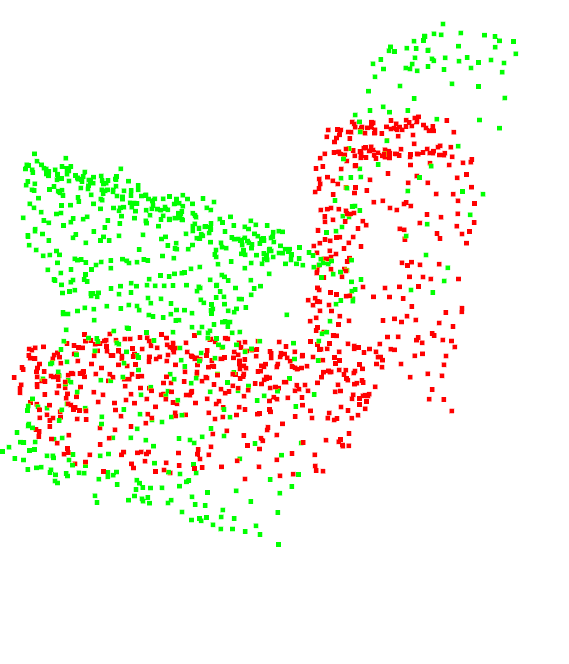} & \includegraphics[width=0.1\textwidth]{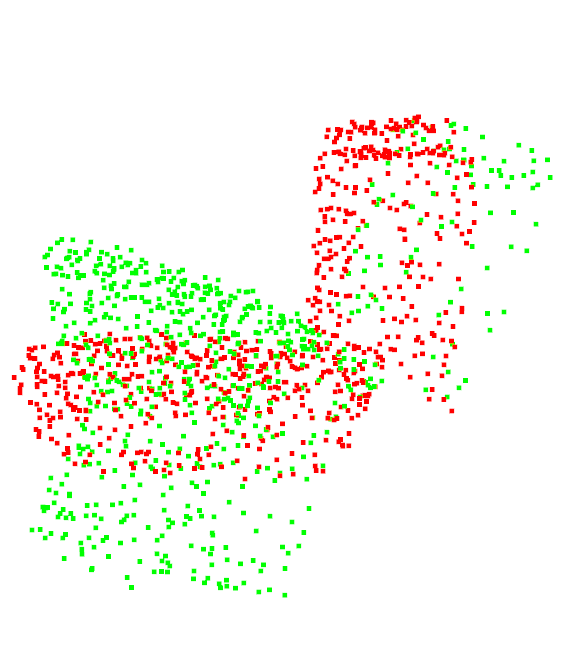} & \includegraphics[width=0.1\textwidth]{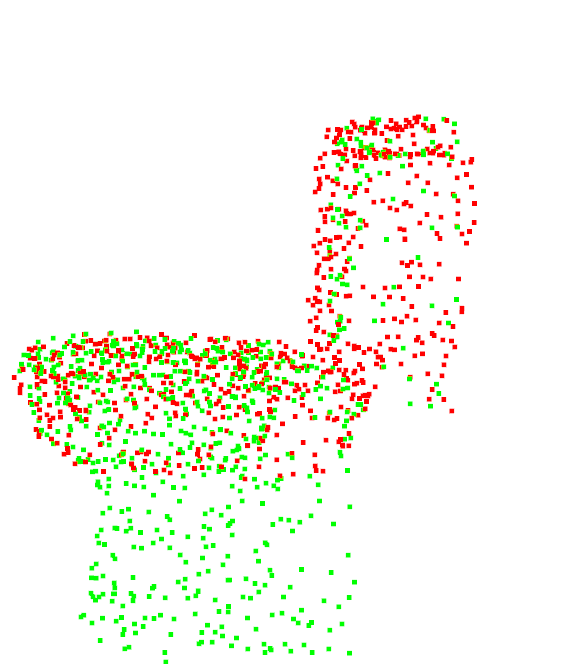} & \includegraphics[width=0.1\textwidth]{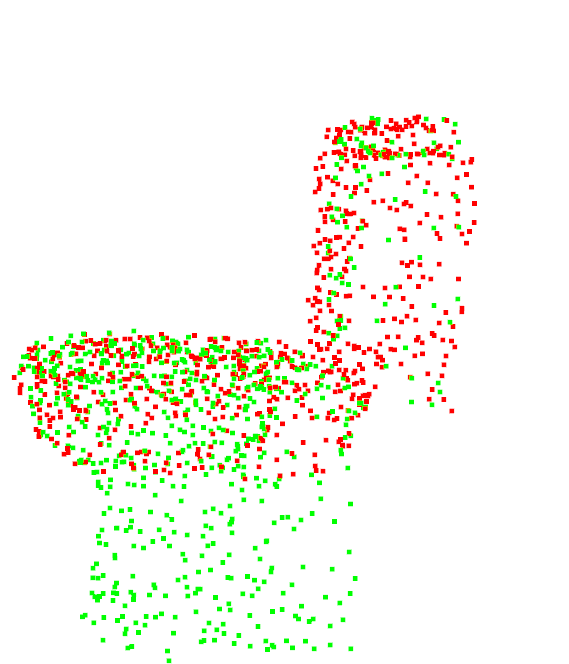} & \includegraphics[width=0.1\textwidth]{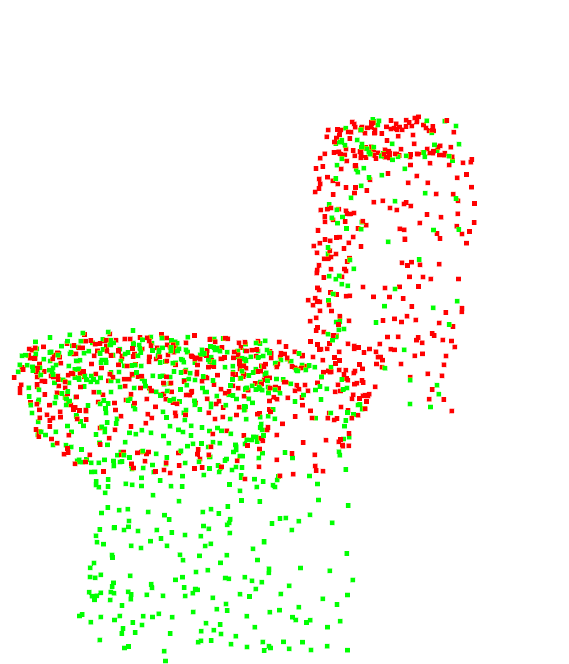} \\
\includegraphics[width=0.1\textwidth]{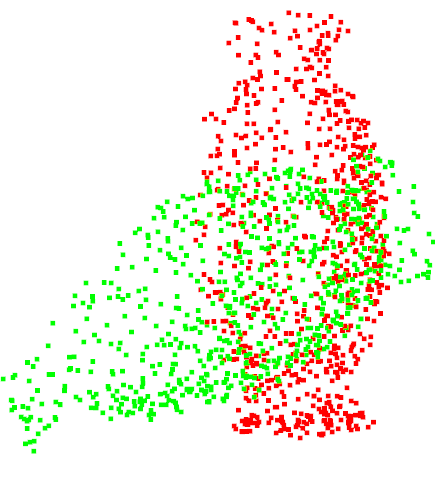} & \includegraphics[width=0.1\textwidth]{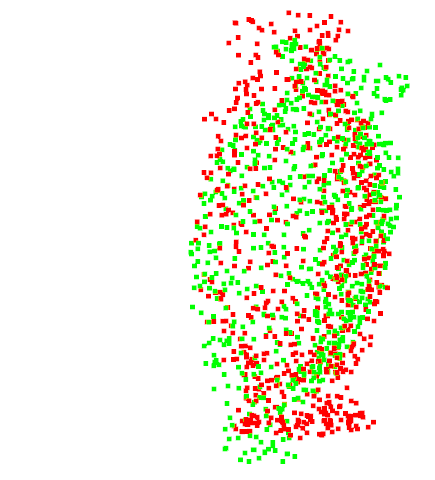} & \includegraphics[width=0.1\textwidth]{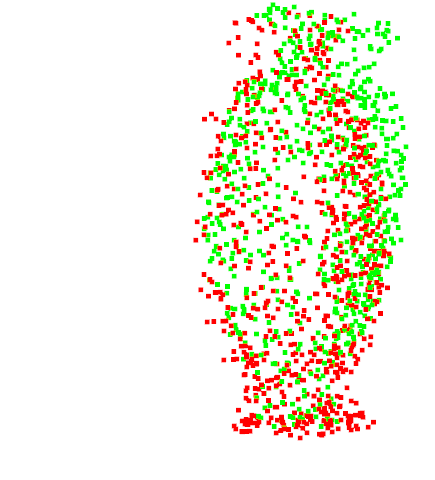} & \includegraphics[width=0.1\textwidth]{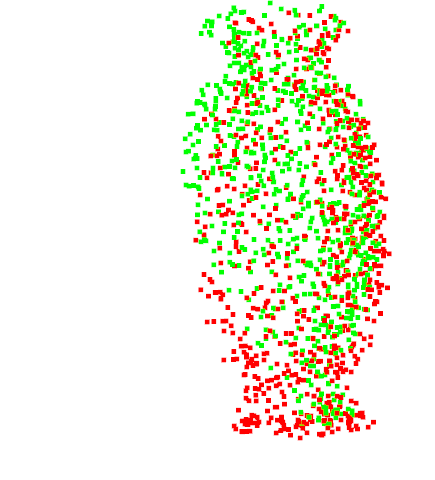} & \includegraphics[width=0.1\textwidth]{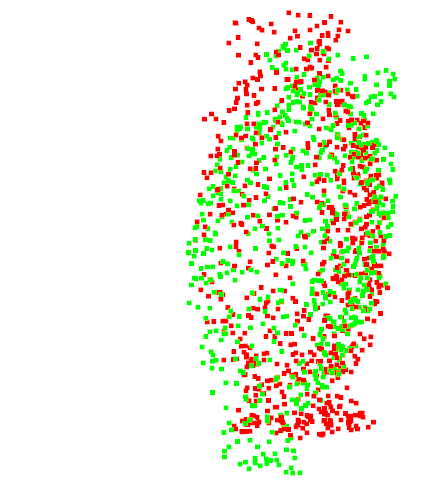} & \includegraphics[width=0.1\textwidth]{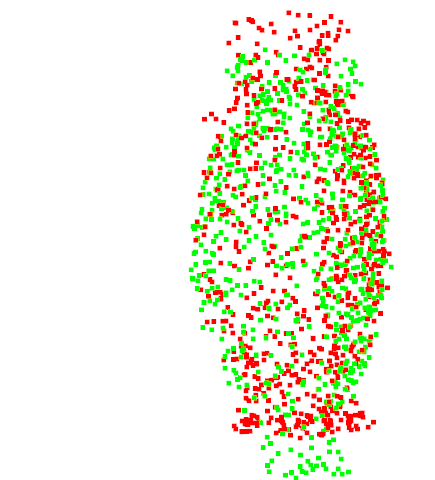} & \includegraphics[width=0.1\textwidth]{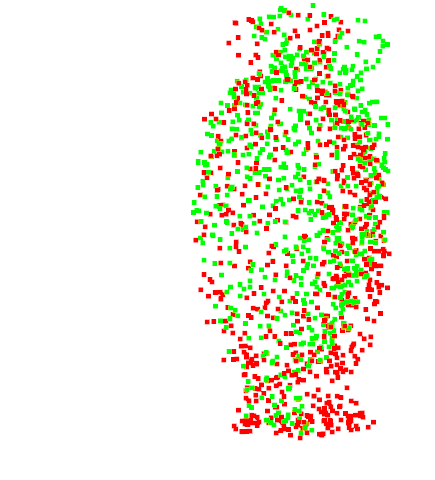} & \includegraphics[width=0.1\textwidth]{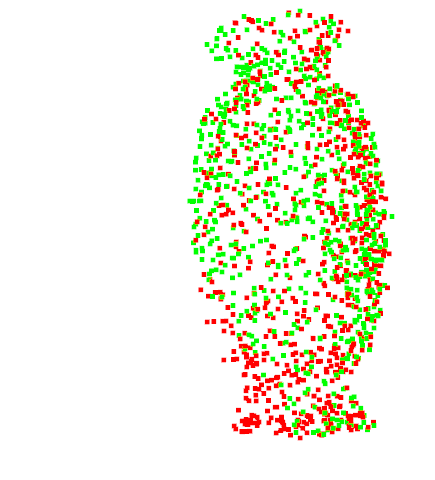} & \includegraphics[width=0.1\textwidth]{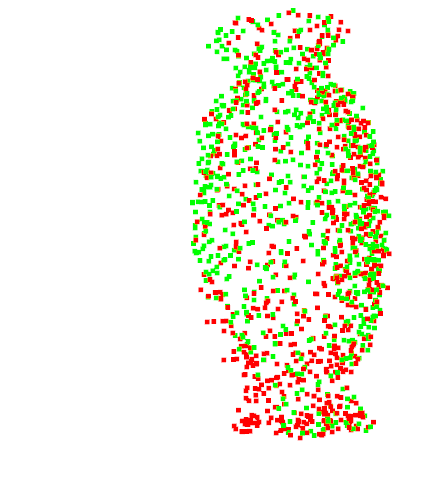} & \includegraphics[width=0.1\textwidth]{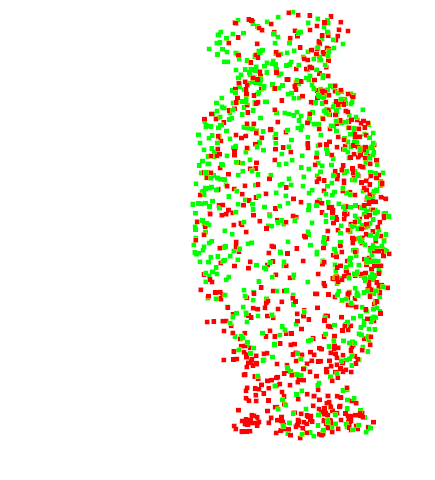} \\
\includegraphics[width=0.1\textwidth]{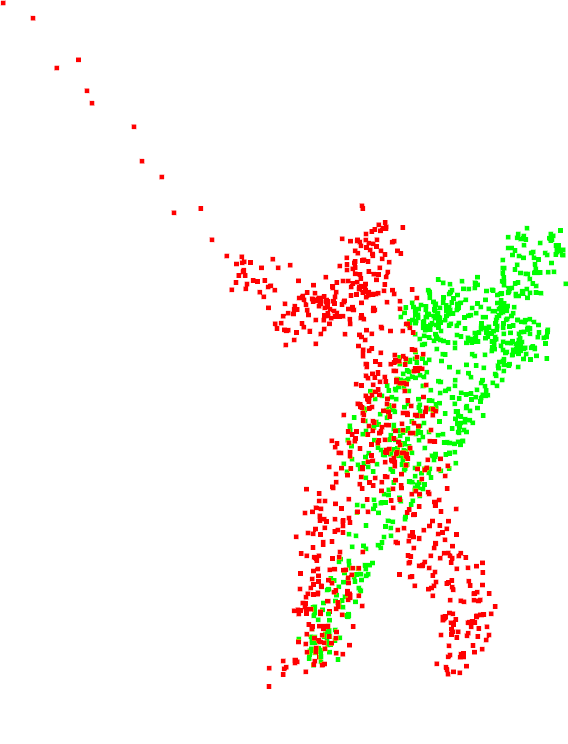} & \includegraphics[width=0.1\textwidth]{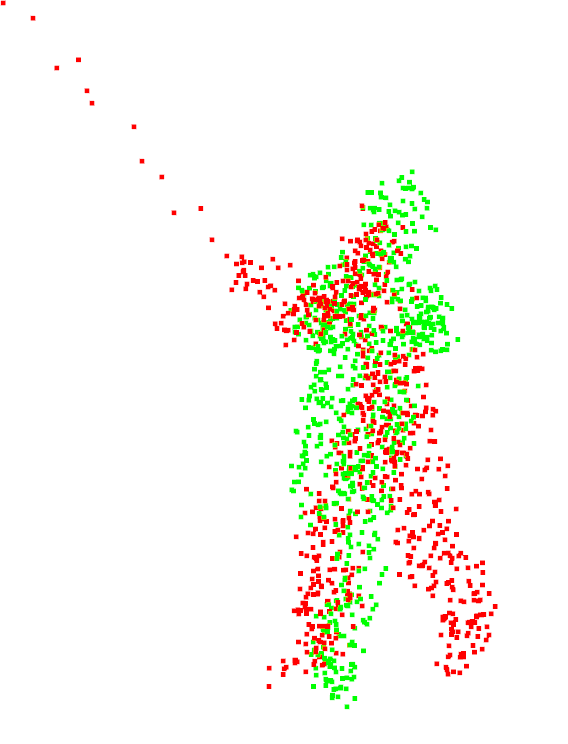} & \includegraphics[width=0.1\textwidth]{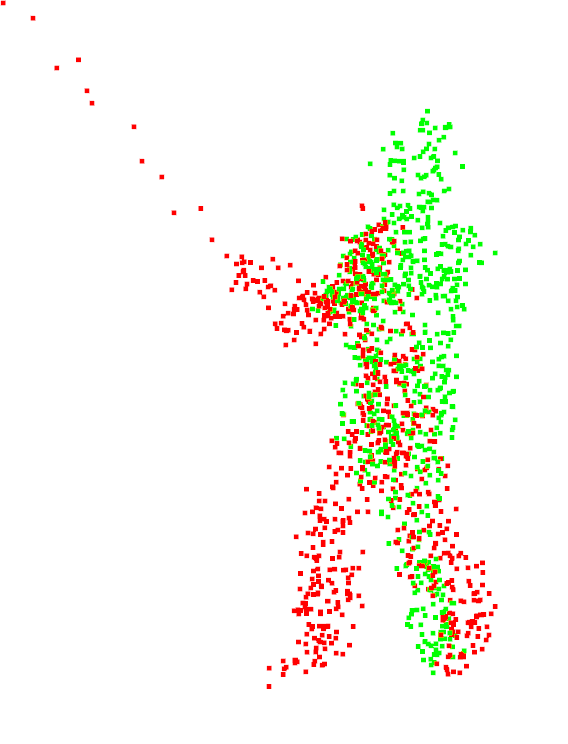} & \includegraphics[width=0.1\textwidth]{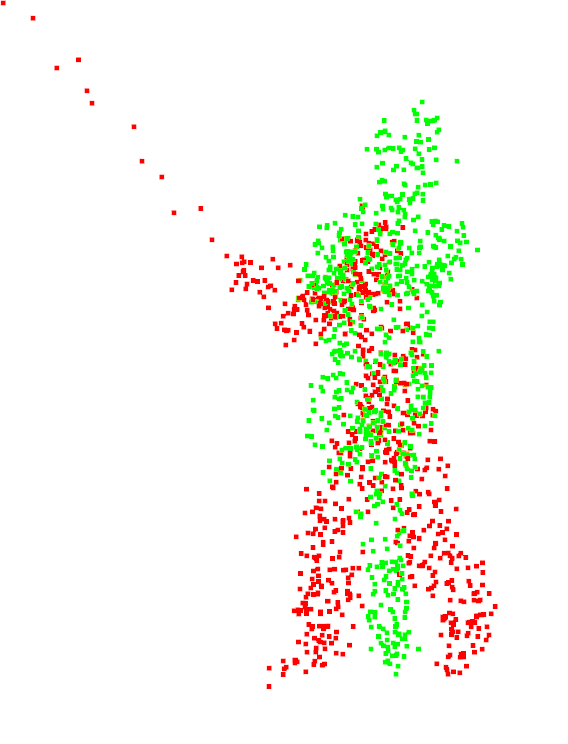} & \includegraphics[width=0.1\textwidth]{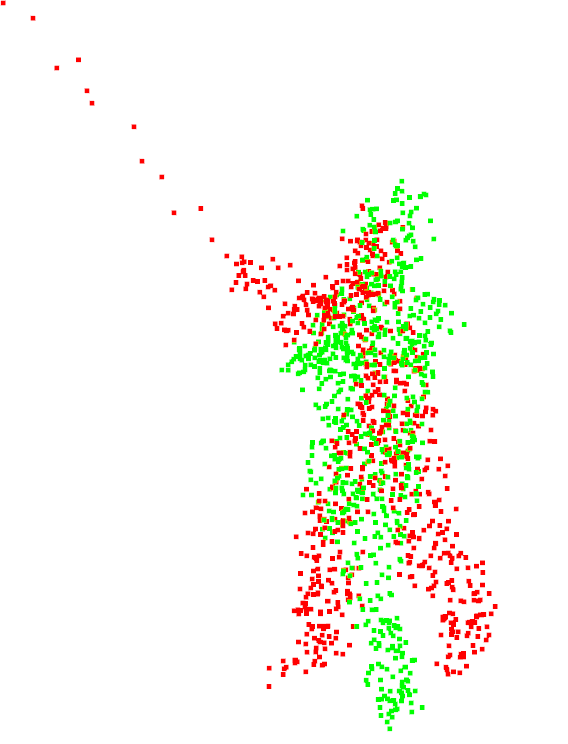} & \includegraphics[width=0.1\textwidth]{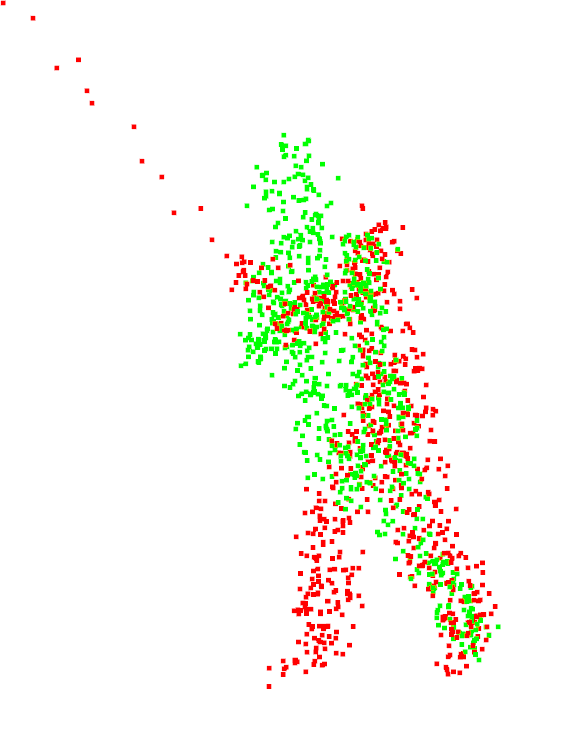} & \includegraphics[width=0.1\textwidth]{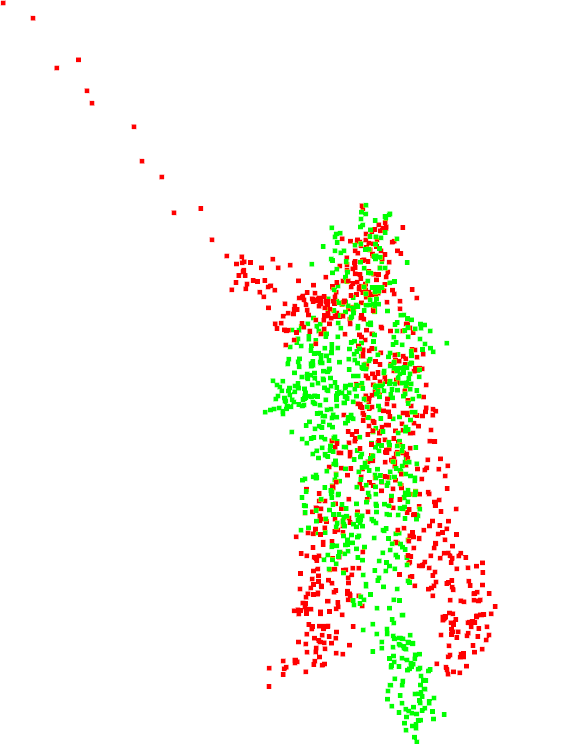} & \includegraphics[width=0.1\textwidth]{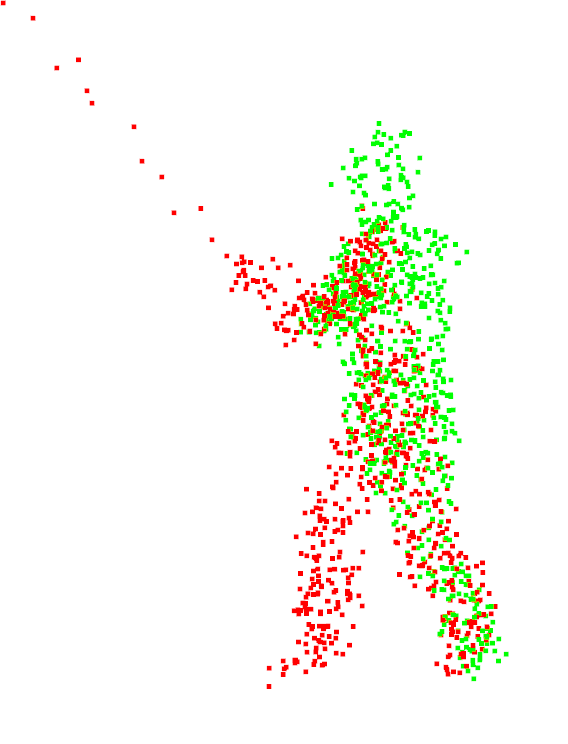} & \includegraphics[width=0.1\textwidth]{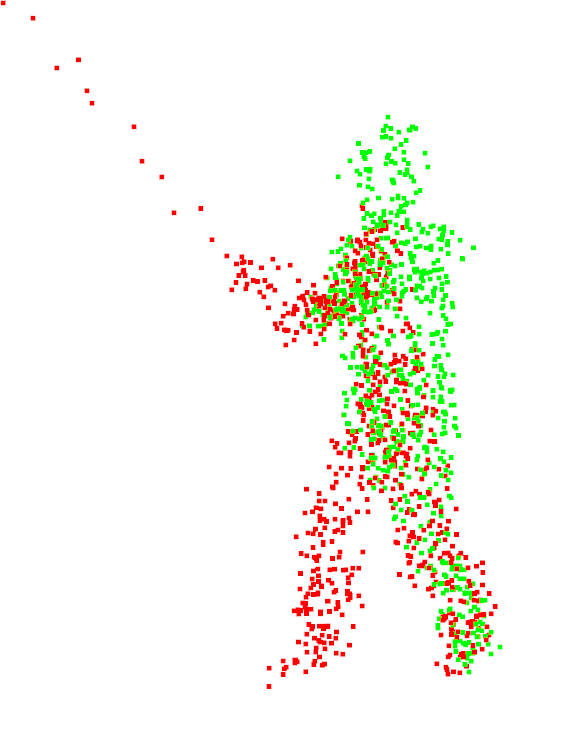} & \includegraphics[width=0.1\textwidth]{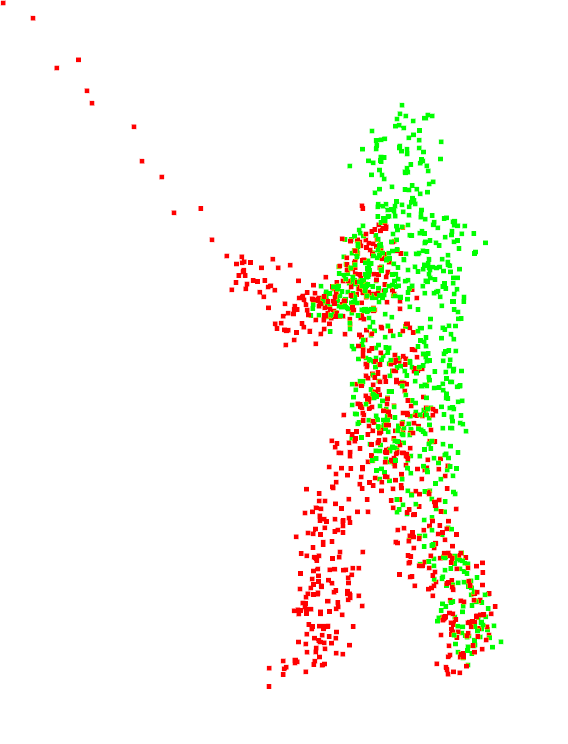} \\
\end{tabular}}
\caption{Registration visualization on ModelNet40 unseen categories with Gaussian noise. The source and target point cloud $X, Y$ are marked in green and red respectively}
\label{fig:reg_vis}
\end{figure}

{\bf Robustness evaluation} We also evaluate the robustness of models in the presence of noise. The noise is sampled from $N(0, 0.01^2)$ and clipped to $[-0.5, 0.5]$. As shown in \autoref{table:3}, FGR is sensitive to noise, and registration error grows sharply. On the contrary, other methods are robust to noise and obtain similar registration error. In addition, our method is still the best among all the methods with $Error(R)$ 1.1566, $Error(t)$ 0.0137 in AO data and $Error(R)$ 1.4656, $Error(t)$ 0.0145 in TO data. The registration visualization is illustrated in \autoref{fig:reg_vis}.

\subsection{Ablation study}

{\bf Overlap prediction} \autoref{table:4} (left) shows the precision and recall of the predicted overlapping points in $X$. The numbers in brackets indicate the threshold $d$ defining in \autoref{eq:1}. The CG module obtains 0.810 and 0.939 in precision with the threshold 0.05 and 0.07, respectively, demonstrating that most of the predicted overlapping points by CG module are positive or near positive overlapping points. With TFMR, ROPNet obtains better precision with 0.913 and 0.987 in the threshold of 0.05 and 0.07. The results reported in TO data showed the overlap prediction is also applicable to symmetric objects. Though OR drops with TFMR, some overlapping points removed in $X$ are with poor descriptor introduced in \autoref{subsec:tfmr}.
\begin{table}
\begin{center}
\scalebox{0.65}{
%\floatrow
%
\begin{tabular}{c|cc|cc}
\hline
 & \multicolumn{2}{|c}{AO} &  \multicolumn{2}{c}{TO} \\
\hline
 & OP (0.05, 0.07) & OR (0.05, 0.07) & OP (0.05, 0.07) & OR (0.05, 0.07) \\
\hline
CG & 0.810 / 0.939 & {\bf 0.836} / {\bf 0.822} & 0.806 / 0.936 & {\bf 0.838} / {\bf 0.824} \\
CG + TFMR & {\bf 0.913} / {\bf 0.987} & 0.382 / 0.347 & {\bf 0.908} / {\bf 0.984} & 0.382 / 0.348 \\
\hline
\end{tabular}
\quad \quad
\begin{tabular}{cccc|cccc}
\hline
Correspondence & Global & Overlap & FMR  & TO $Error(R)$ & TO $Error(t)$ \\
\hline
$\surd$ & & & & 5.9518 & 0.0843 \\
$\surd$ & $\surd$ & & & 4.8219 & 0.0502 \\
$\surd$ & & $ \surd$ & & 3.3361 & 0.0378  \\
$\surd$ & $\surd$ & $\surd$ & & 1.9592 & 0.0189 \\
$\surd$ & $\surd$ & $\surd$ & $\surd$ & {\bf 1.4656} & {\bf 0.0145} \\
\hline
\end{tabular}
}
\end{center}
\caption{Left: Overlap Precision (OP) and Recall (OR). Right: Ablation studies of different components for registration error. They are both on ModelNet40 unseen categories with Gaussian noise.}
\label{table:4}
\end{table}

{\bf Registration error} As shown in \autoref{table:4} (right), we regard the first row as the benchmark, which just learns point features and generates similarity matrix for registration. The second row and the third row demonstrate global context features for initialized alignment, and the overlap module can both reduce the rotation error and translation error. The overlap module boosts registration accuracy more. With a combination of coarse initialization and the overlap module, as shown in the fourth row, $Error(R)$, $Error(t)$ are reduced significantly to 1.9592, 0.0189. Furthermore, with FMR module, we gain the lowest error with $Error(R)$ 1.4656 and $Error(t)$ 0.0145.

\section{Conclusion}

We proposed ROPNet, a new method for partially overlapping point clouds registration. A context-guided (CG) module is introduced for initial alignment and overlapping points prediction. Based on the CG module, we included a TFMR module to extract point features based on all input points, obtain representative overlapping points, and transform partial-to-partial registration into partial-to-complete registration. Experiment results showed that ROPNet obtained the lowest error with robustness to noise and generalization ability to unseen shape categories while keeping efficiency. Future work should be considered to refine the design of the proposed network to operate on large-scale point clouds and expand its generalization to real-world scanned data.

\bibliographystyle{unsrt}  
\bibliography{references}

\clearpage

{\large{\bf{\raggedright}{Appendix}}}

% {\large{\bf{\raggedright}{A. Architecture}}}
\section*{A. Architecture}

\begin{figure}[t]
\centering
\includegraphics[width=0.98\linewidth]{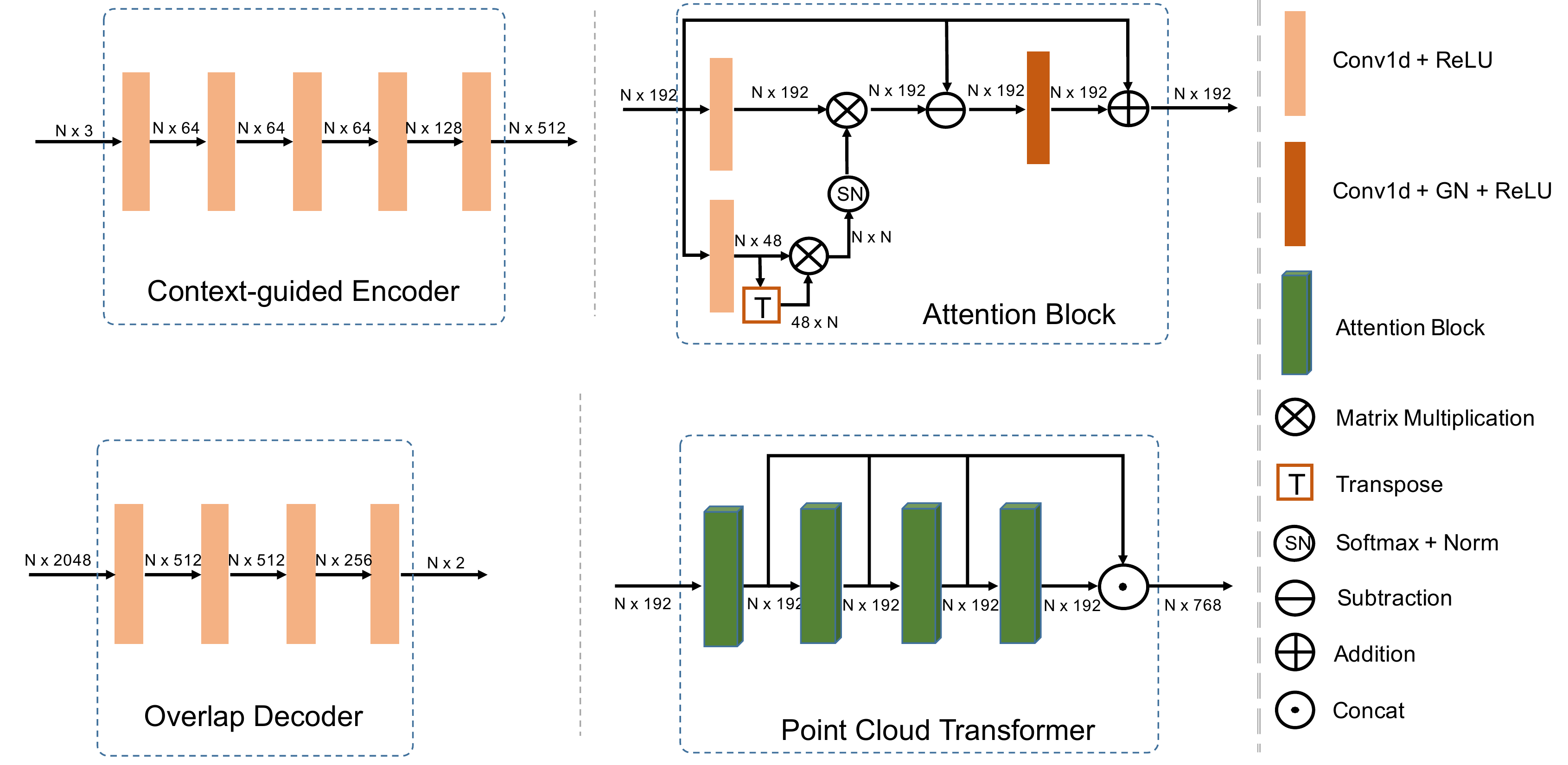}
\caption{Left-top: Architecture of encoder of context-guided module. Left-bottom:  Architecture of overlap decoder. Right-top: Attention block for point cloud Transformer. Right-bottom: Architecture of point cloud Transformer.}
\label{fig:S_architecture}
\end{figure}

\paragraph{Context-guided encoder}
The encoder of the context-guided (CG) module is a PointNet without T-Net for transformations as shown in \autoref{fig:S_architecture} (left-top). It contains 5 Conv1d (64, 64, 64, 128, 512) and ReLU layers without normalization operations. It takes the point cloud $X$ as input and outputs a 512-dimensional feature $F_X$ for each point.

\paragraph{Initialization decoder}
The initial alignment decoder is a MLP that contains 4 FC (512, 512, 256, 7) and ReLU layers without normalization operations. It takes the fused global features $[F^g_X; F^g_Y]$ as input and outputs a 7-dimensional vector $v$ representing the rotation and translation transformation

\paragraph{Overlap decoder}
The overlap decoder is a PointNet without T-Net for transformations as shown in \autoref{fig:S_architecture} (left-bottom). It contains 4 Conv1d (512, 512, 256, 2) and ReLU(except the last layer) layers without normalization operations. It takes the features output from the Information Interactive module and outputs an overlap score for each point.

\paragraph{Point cloud Transformer}

As shown in \autoref{fig:S_architecture} (right-bottom), the point cloud Transformer contains four attention blocks and concatenates outputs from each attention block as the final output features. As shown in \autoref{fig:S_architecture} (right-top), attention block is a offset attention which calculates the difference between the self-attention features and the input features. Each attention block outputs a 192-dimensional feature. Thus the output of the point cloud Transformer is 768-dimensional features. We use GN instead of BN in point cloud Transformer.

\paragraph{Loss function}
The total loss is the weighted summation of the three losses:
\begin{equation}
L_{total} = \alpha \cdot L_{fin} + \beta  \cdot L_{ol} + \lambda \cdot L_{init},
\end{equation}
where $\alpha$, $\beta$ and $\gamma$ are set to 1, 0.1 and 1, respectively.
%FGF: which values they have ?

% {\large{\bf{\raggedright}{B. Implementation Details}}}
\section*{B. Implementation Details}

The number of reserved points ($\text{top-}N_1$) is 448 after removal operation based on overlap score. $\text{Top-}prob$ is set to 0.6 and 0.4 for training and test, respectively. We set $\text{top-}k$ to 3 and 1 for training and test, respectively.  The overlap threshold $d$ is set to 0.05. We train ROPNet in a non-iterative manner. However, we run 2 iterations for the TFMR module during test. For the ModelNet40 dataset, we train for 600 epochs, using Adam with an initial learning rate of 0.0001. The learning rate changes using a cosine annealing schedule with the first restart set to 40 iterations and an increasing factor of 2. We use batch size 8 in experiments. Data augmentation includes horizontal flip and adding Gaussian noise with $\sigma=0.5$

% {\large{\bf{\raggedright}{C. Experiments}}}
\section*{C. Experiments}

% {\large{\bf{\raggedright}{C.1 Benchmarks implementation details}}}
\subsection*{C.1 Benchmarks implementation details}

We compare our method with traditional and learning-based methods, including ICP, FGR, PCRNet, DCP, IDAM, DeepGMR, and RPMNet. For a fair comparison, we evaluate them on the same test set (the same input point clouds and transformation matrix) and retrain them with the same generation strategy for point cloud pairs, randomly. ICP and FGR are implemented using the Open3D library. PCRNet and DeepGMR are designed for complete-to-complete registration. The proposed loss is invalid for partial-to-partial registration, so we modified the loss as conducted in RPMNet for effective training. It is also noted that the hand-crafted RRI features are not essential for DeepGMR, and we found that the training is not converging with RRI features as input, so we remove RRI features in both the training and testing stage. For DCP, we adopt DCP-v2 with an attention model and SVD. For IDAM, we train with both GNN-based features and FPFH-based features and find that they both overfit the training set, so we reported the results of IDAM-GNN model, which are better in performance. PRNet may be the first learning-based model to solve partial-to-partial registration. However, the code released by the authors is prone to collapse when training, so we don't use it for comparison in this work. 

% {\large{\bf{\raggedright}{C.2 Ablation study}}}
\subsection*{C.2 Ablation study}

The following experiments in this subsection are all conducted on the ModelNet40 unseen categories with Gaussian noise.

\begin{figure}[t]
\centering
\includegraphics[width=0.49\linewidth]{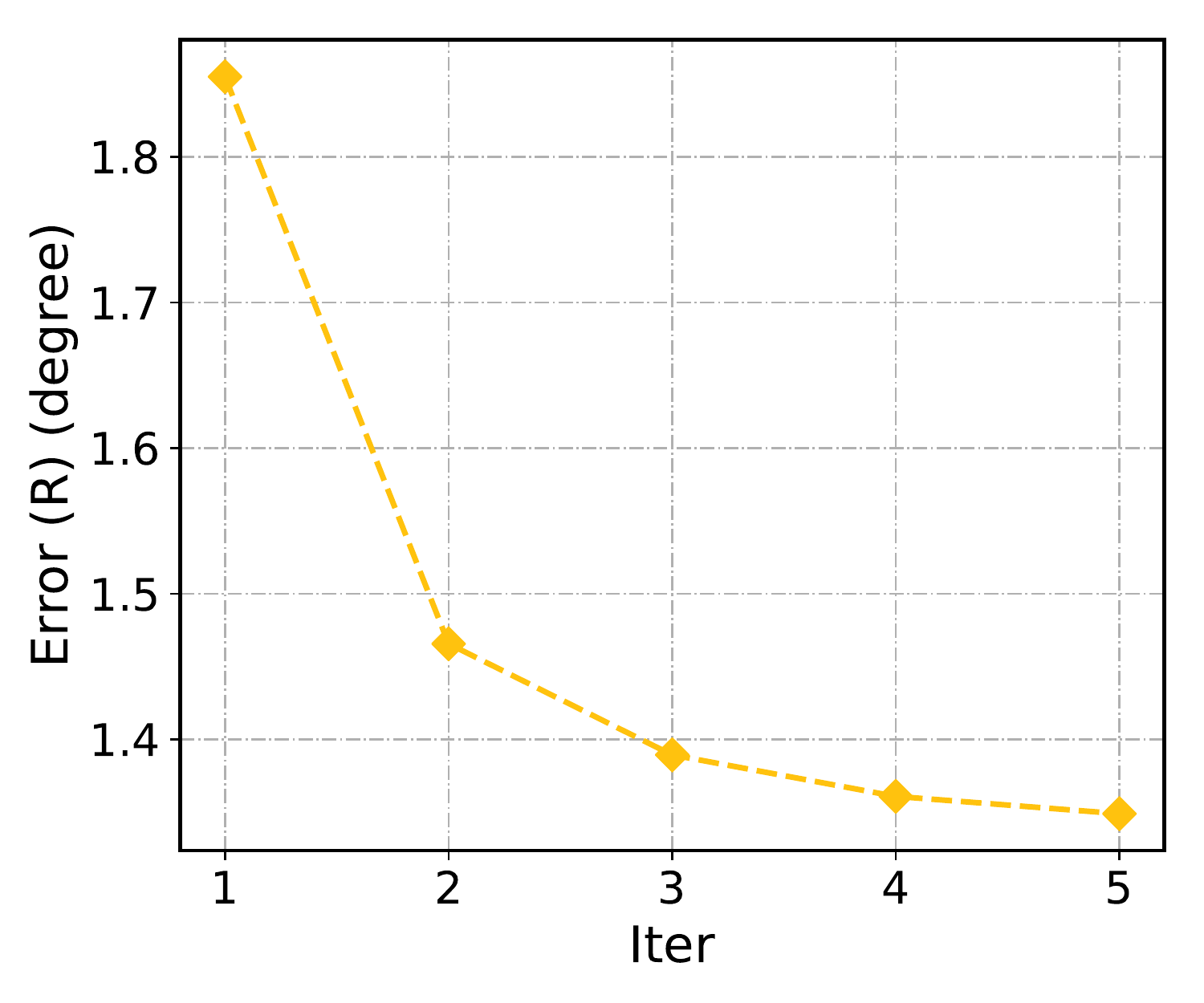}
\hfill
\includegraphics[width=0.49\linewidth]{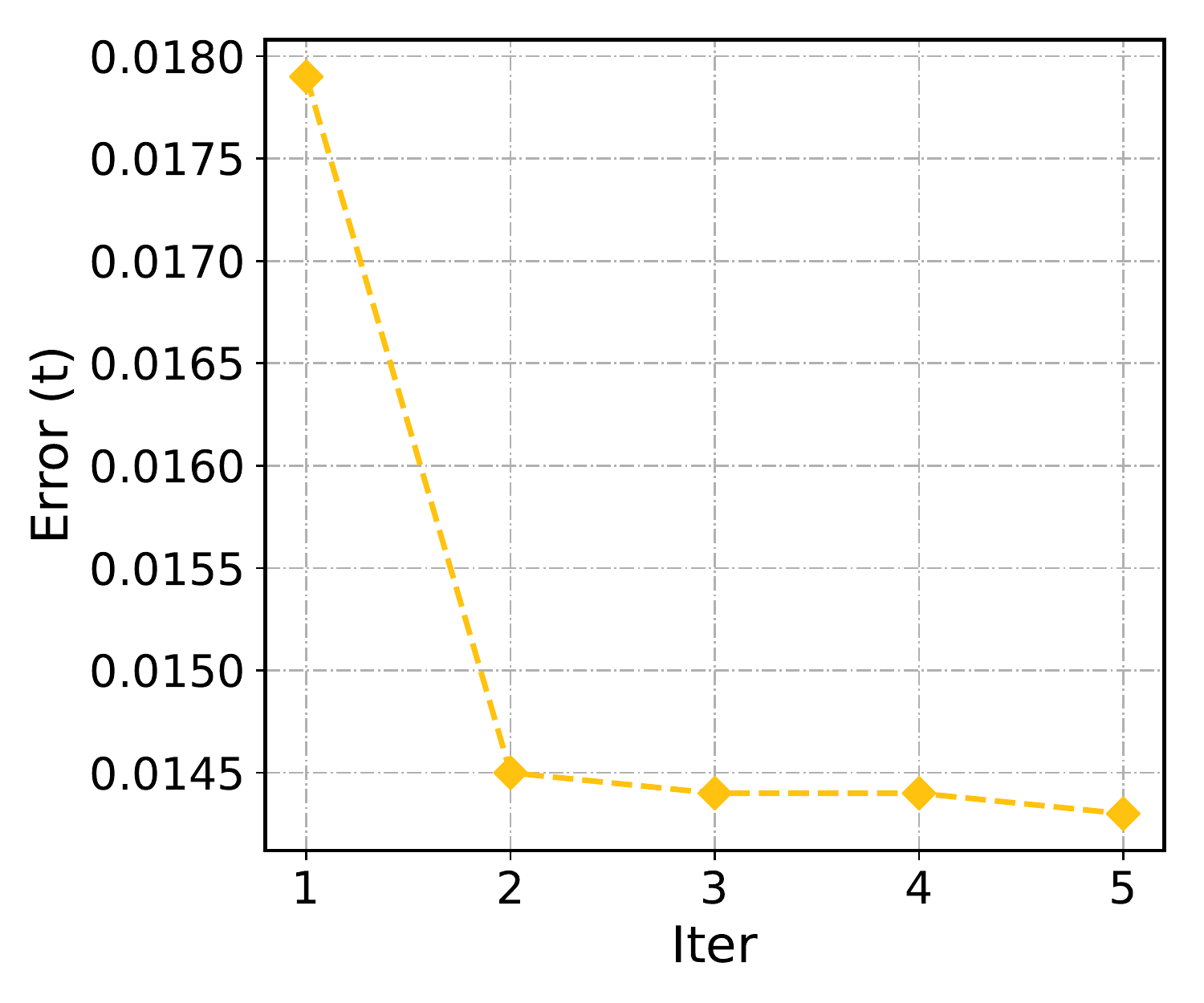}
\caption{Left: $Error(R)$ with different iterations. Right: $Error(t)$ with different iterations.}
\label{fig:iter}
\end{figure}

\paragraph{Number of iterations}
We evaluated $Error(R)$ and $Error(t)$ based on different iterations. The results are shown in  \autoref{fig:iter}. As the number of iterations increases, the error decreases. The error drops sharply in the second iteration. Considering the running time, we only iterate two times in our ROPNet. 

\begin{figure}[t]
\centering
\includegraphics[width=0.49\linewidth]{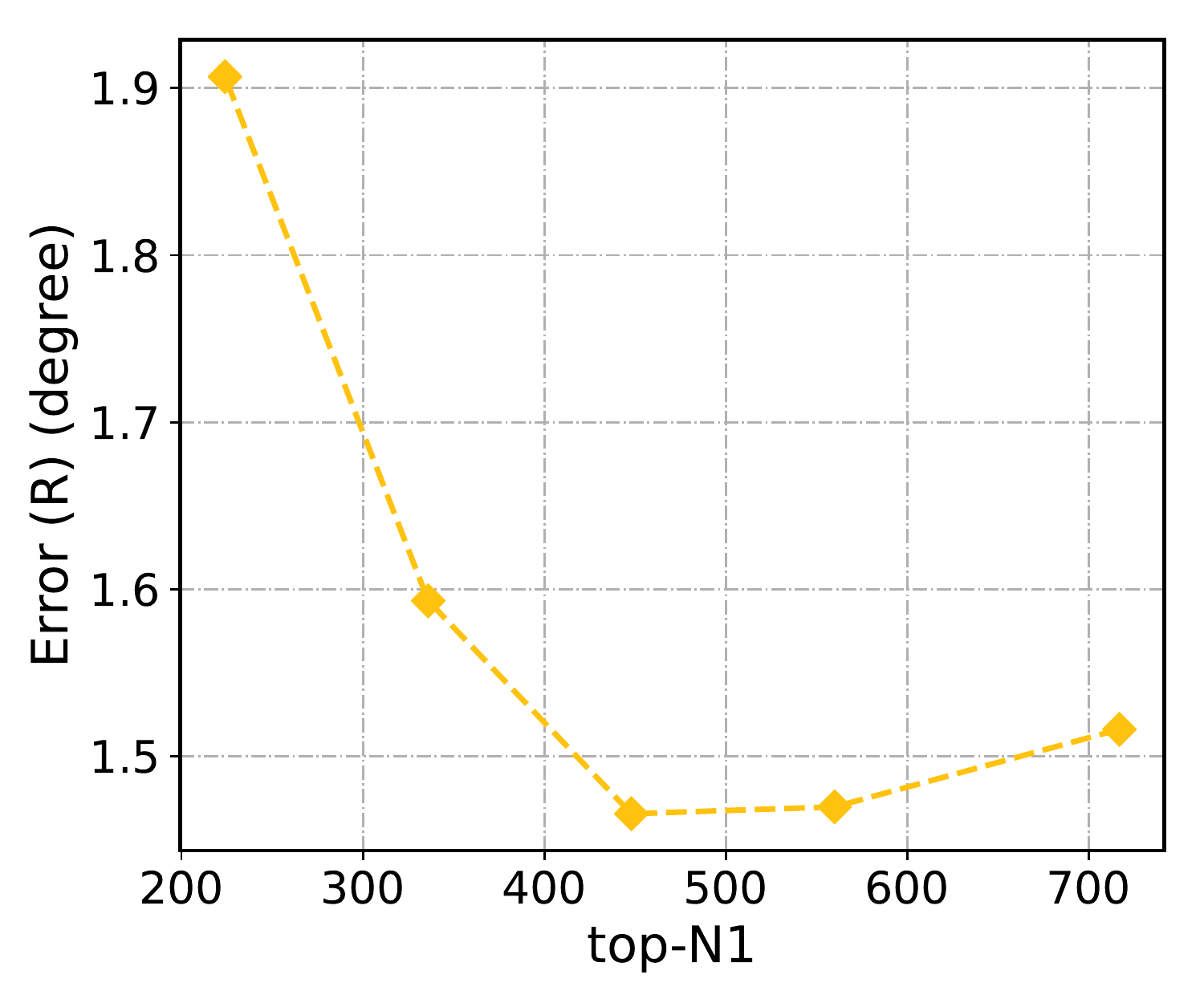}
\hfill
\includegraphics[width=0.49\linewidth]{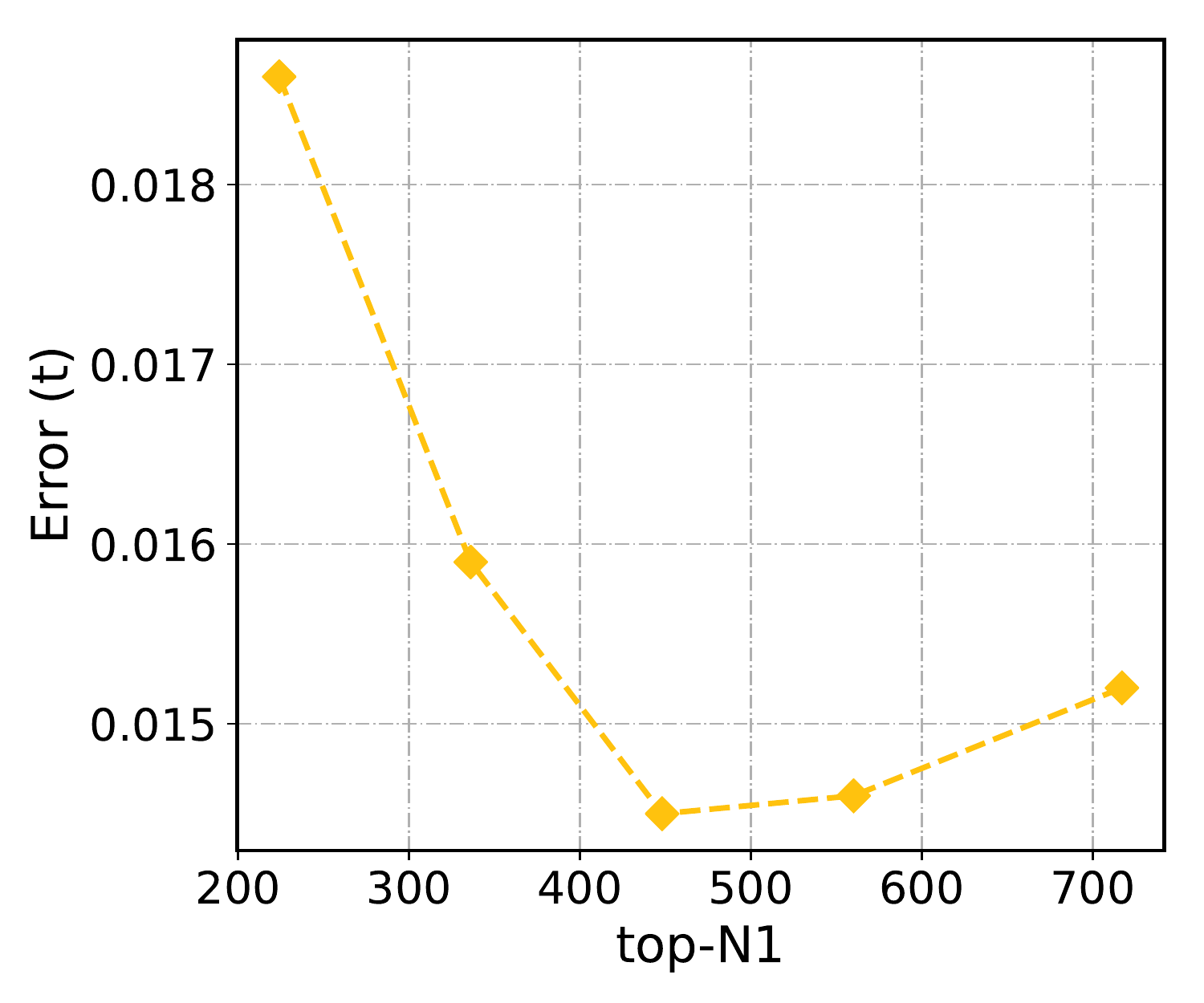}
\caption{Left: $Error(R)$ with different $N_1$. Right: $Error(t)$ with different $N_1$.}
\label{fig:top-n1}
\end{figure}

\paragraph{Top-N1} Based on overlap score, we keep $\text{top-}N_1$ points in source point cloud $X$ for registration. As shown in \autoref{fig:top-n1}, we evaluated $Error(R)$ and $Error(t)$  for different $N1 \in \{717, 560, 448, 336, 224 \}$. We obtained sub-optimal registration results based on all points (717) for registration, because some points are non-overlapping points. We obtained the lowest error based on 448 points for registration.

\begin{figure}[t]
\centering
\includegraphics[width=0.49\linewidth]{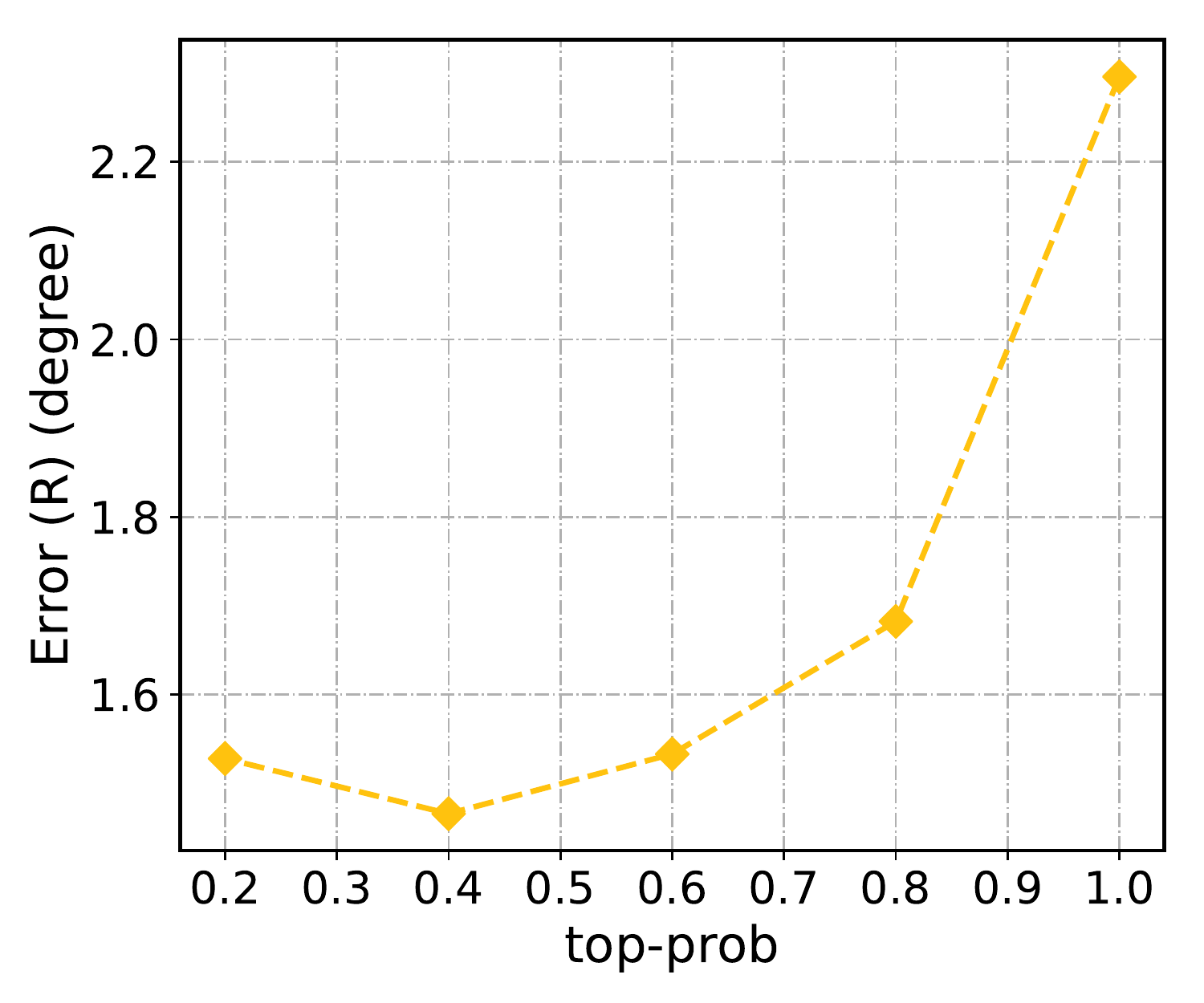}
\hfill
\includegraphics[width=0.49\linewidth]{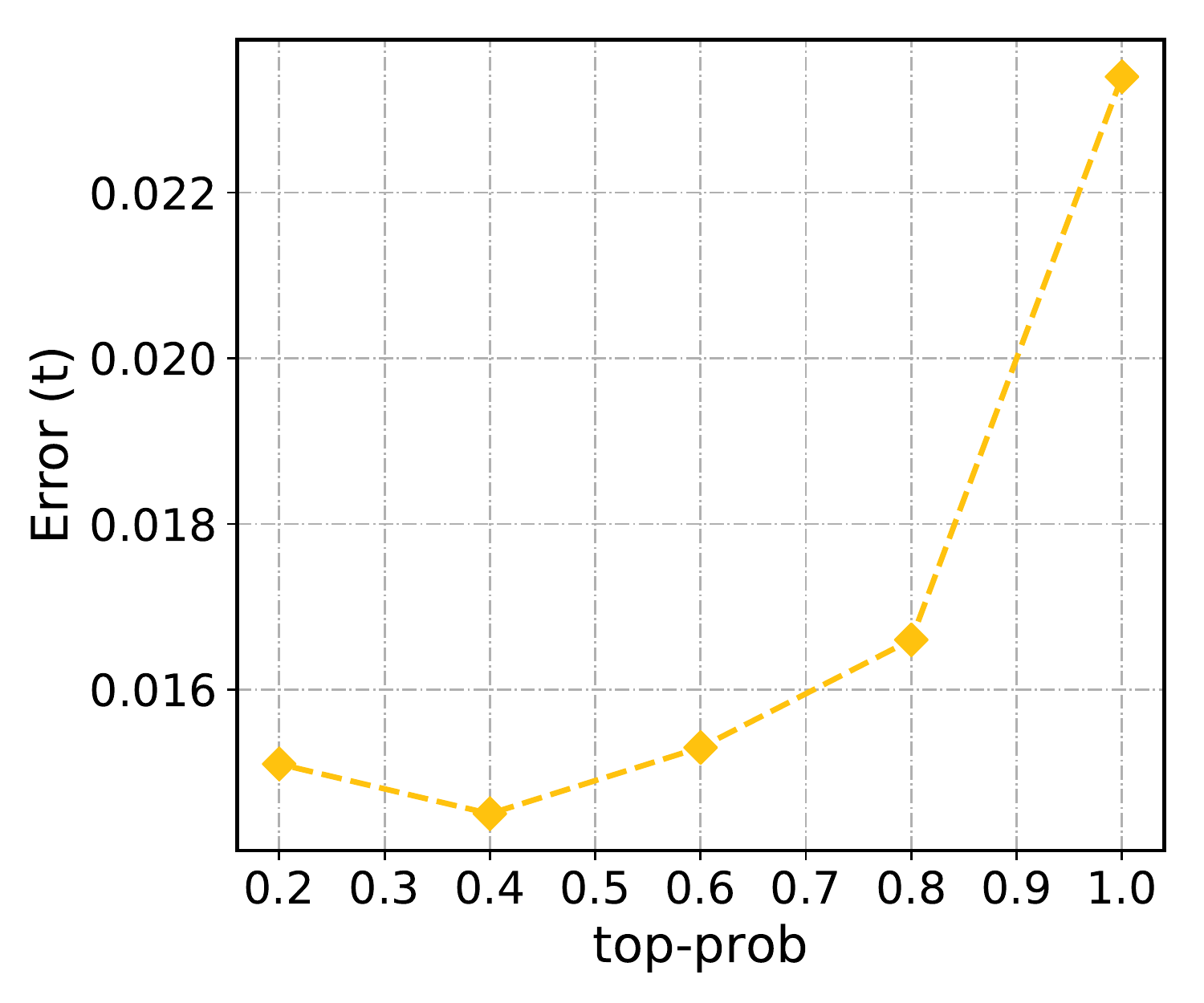}
\caption{Left: $Error(R)$ with different $\text{top-}prob$. Right: $Error(t)$ with different $\text{top-}prob$.}
\label{fig:top-prob}
\end{figure}

\paragraph{Top-prob} Based on feature matching, we keep $\text{top-}prob$ points for registration whose features are descriptive. We evaluated $Error(R)$ and $Error(t)$ based on different $\text{top-}prob$ with values: $0.2, 0.4, 0.6, 0.8$ and $1$. As shown in \autoref{fig:top-prob}, all points($\text{top-}prob=1$) for registration is sub-optimal due to selected points are not representative. We obtained the best performance based on $\text{top-}prob=0.4$.

\begin{figure}[t]
\centering
\includegraphics[width=0.49\linewidth]{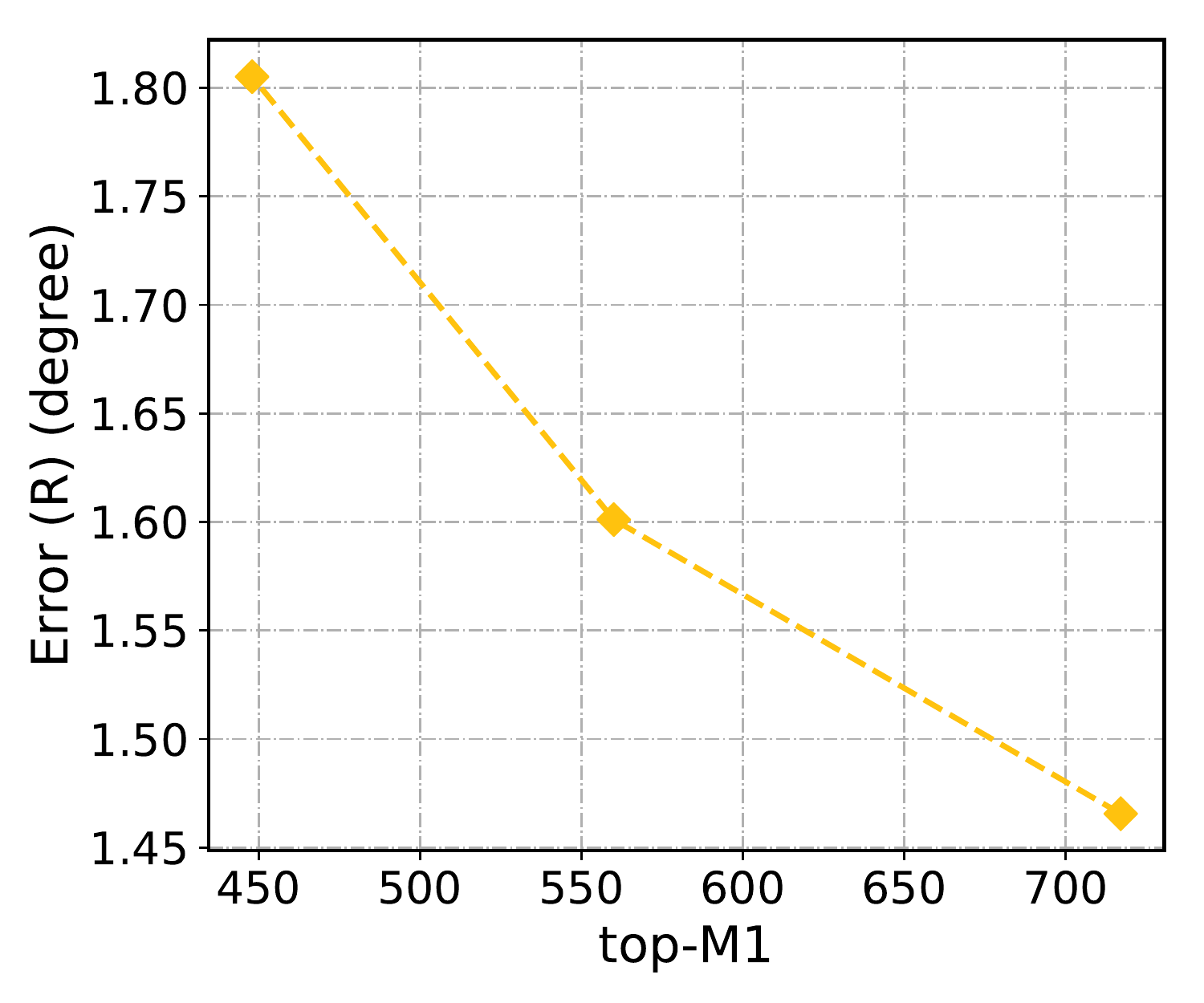}
\hfill
\includegraphics[width=0.49\linewidth]{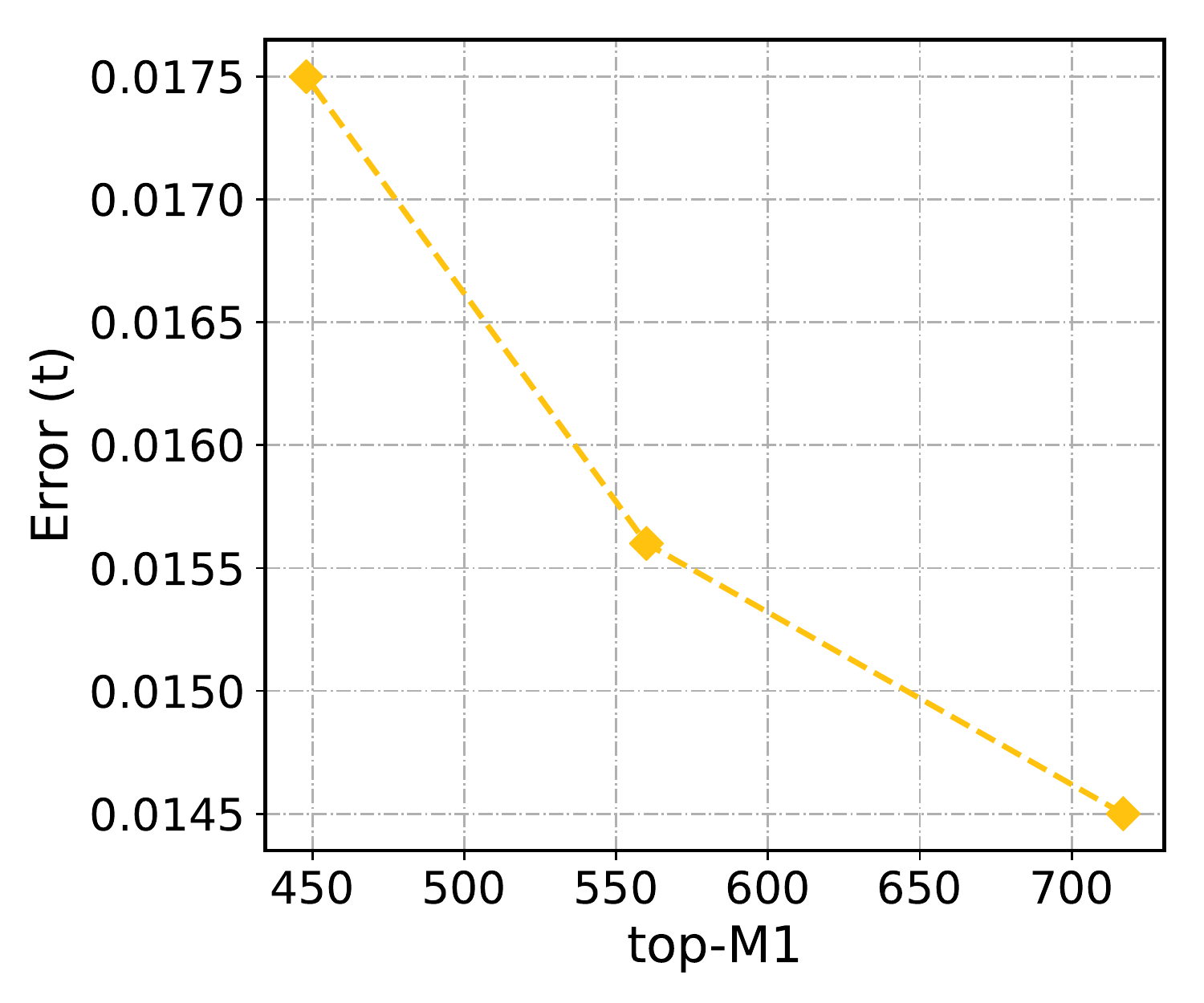}
\caption{Left: $Error(R)$ with different $M_1$. Right: $Error(t)$ with different $M_1$.}
\label{fig:top-m1}
\end{figure}

\paragraph{Top-M1} Based on overlap score, we keep $\text{top-}M_1$ points in target point cloud $Y$ for registration. As shown in \autoref{fig:top-m1}, we evaluate the $Error(R)$ and $Error(t)$ for different values $M1 \in \{ 717, 560, 448\}$. We obtained the lowest error based on all points (717) for registration, demonstrating that some overlapping points, mistakenly removed from the target point cloud, affect the registration.

% {\large{\bf{\raggedright}{C.3 Additional experiments}}}
\subsection*{C.3 Additional experiments}

\begin{figure}
\centering
\scalebox{0.9}{
\begin{tabular}{ccccc}
Inputs & CG & CG + TFMR & G.T. \\
\includegraphics[width=0.24\textwidth]{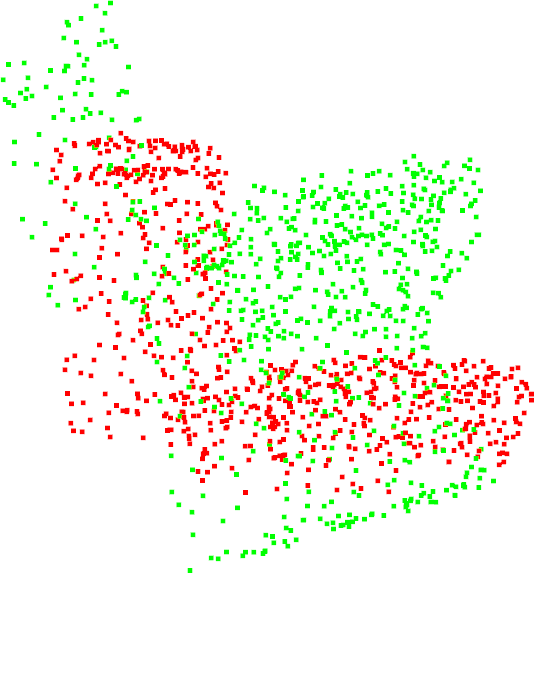} &  \includegraphics[width=0.24\textwidth]{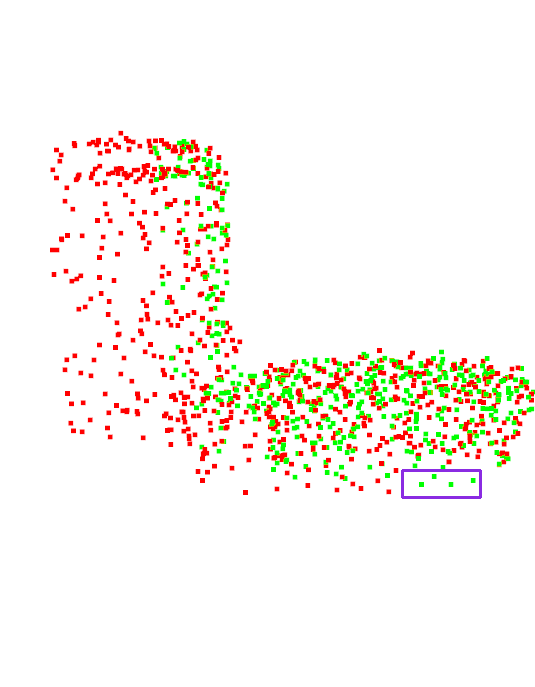} &  \includegraphics[width=0.24\textwidth]{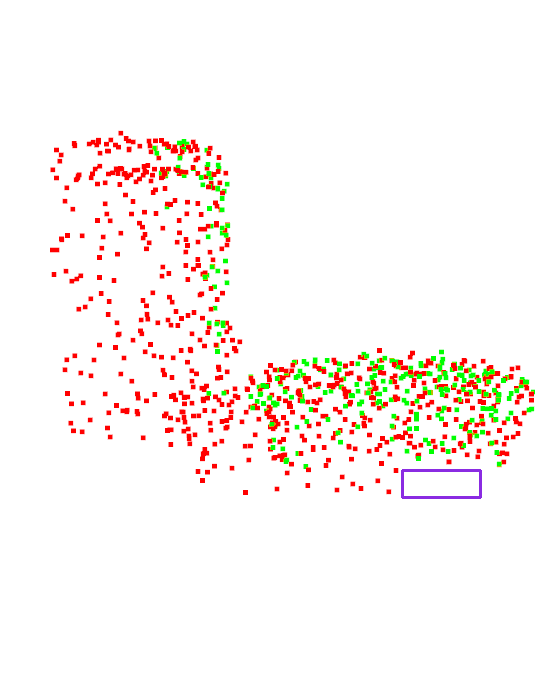} &
\includegraphics[width=0.24\textwidth]{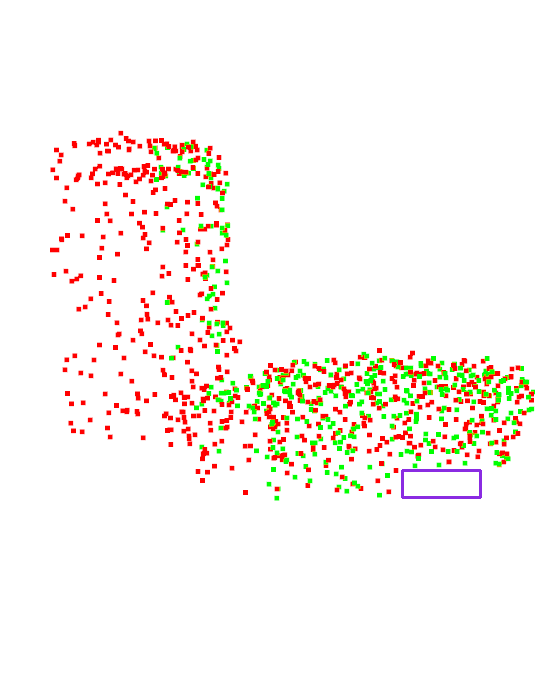} \\
\includegraphics[width=0.24\textwidth]{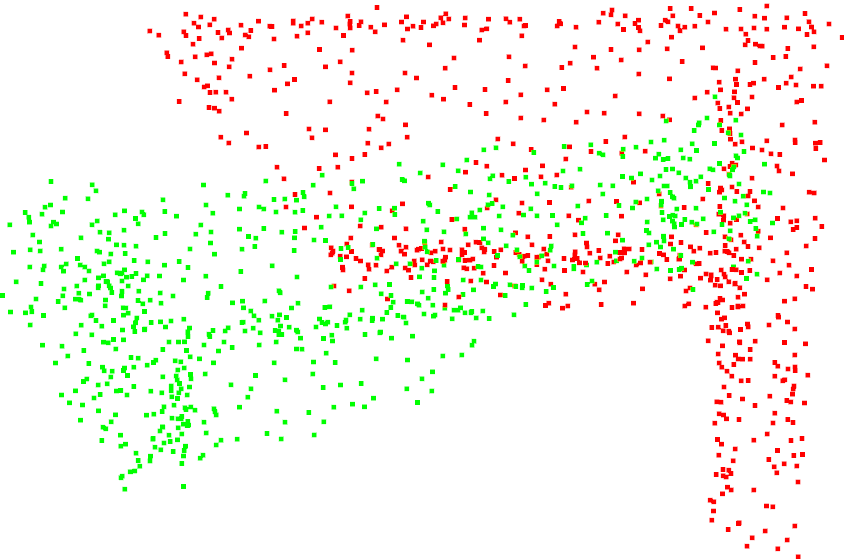} &  \includegraphics[width=0.24\textwidth]{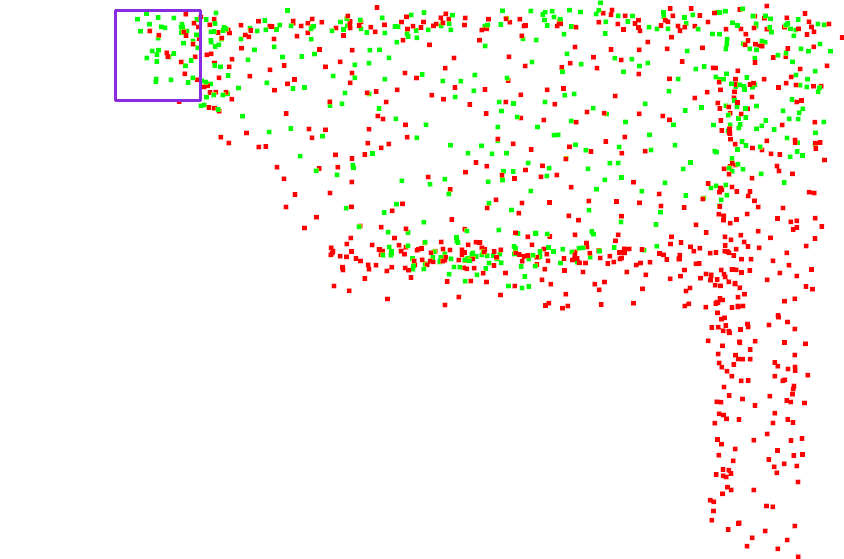} &  \includegraphics[width=0.24\textwidth]{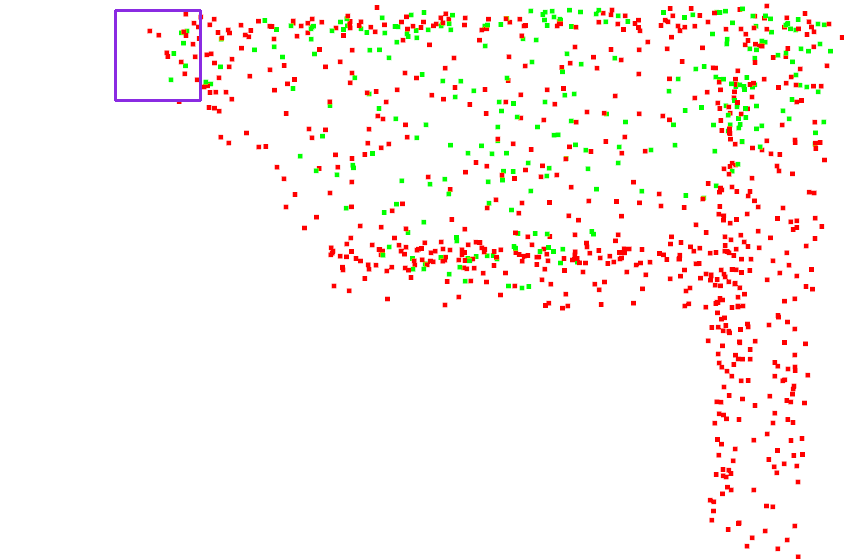} &
\includegraphics[width=0.24\textwidth]{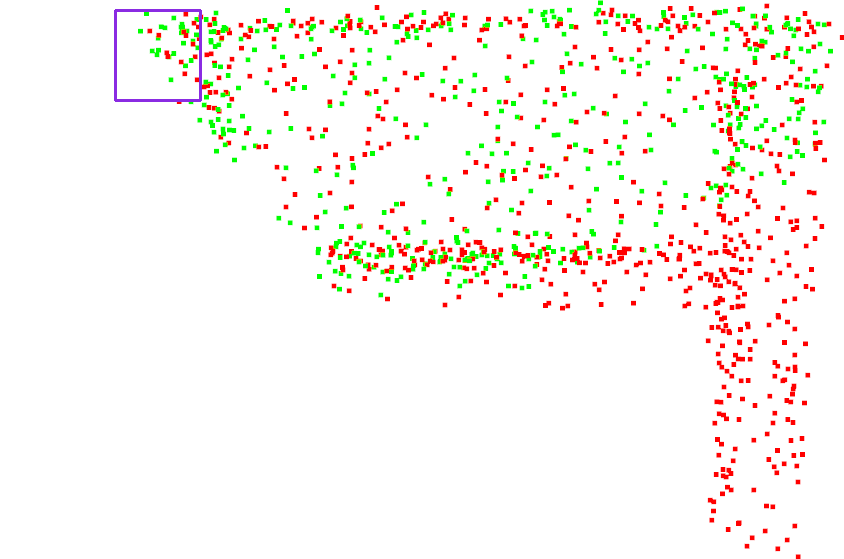} \\
\includegraphics[width=0.24\textwidth]{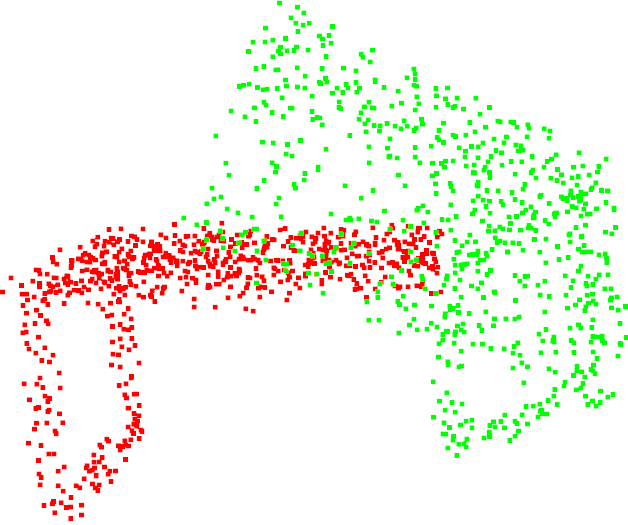} &  \includegraphics[width=0.24\textwidth]{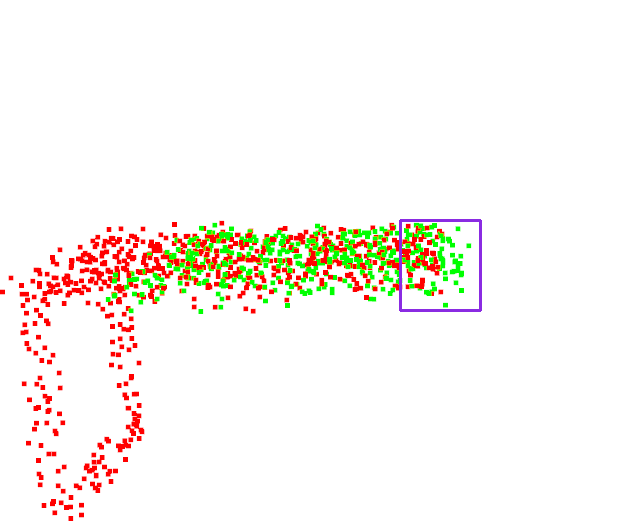} &  \includegraphics[width=0.24\textwidth]{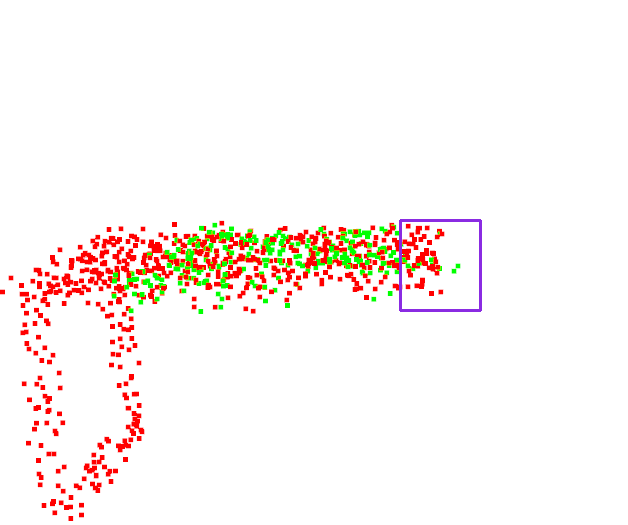} &
\includegraphics[width=0.24\textwidth]{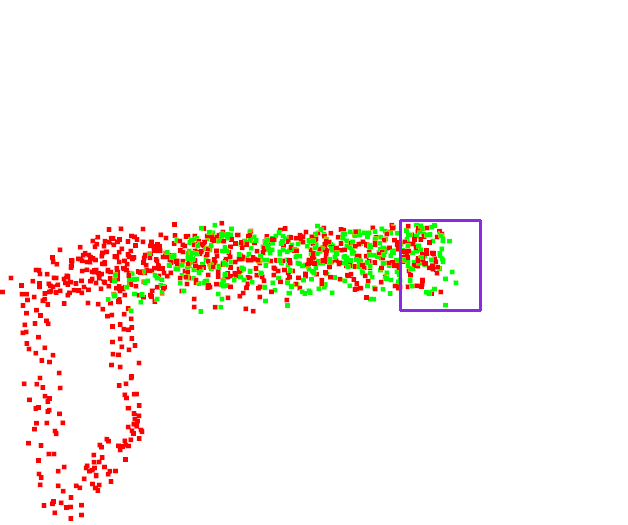} \\
\end{tabular}}
\caption{Predicted representative overlapping points on ModelNet40 unseen categories with Gaussian noise. The source and target point cloud $X, Y$ are colored in green and red, respectively. The second column shows the predicted overlapping points in $X$ by the CG module. As shown in the third column, TFMR further removes non-representative points based on feature matching, including some non-overlapping points depicted in purple rectangles. The last column shows the ground truth overlapping points in $X$.}
\label{fig:reg_ol}
\end{figure}

\paragraph{Representative overlapping points} Here, we visualize the representative overlapping points on ModelNet40 unseen categories with Gaussian noise. As shown in \autoref{fig:reg_ol}, the CG module predicts the overlapping (or nearly overlapping) points. The TFMR module removes the non-representative points (including some non-overlapping points shown in the purple rectangles) based on feature matching.

\begin{figure}
\centering
\includegraphics[width=0.98\linewidth]{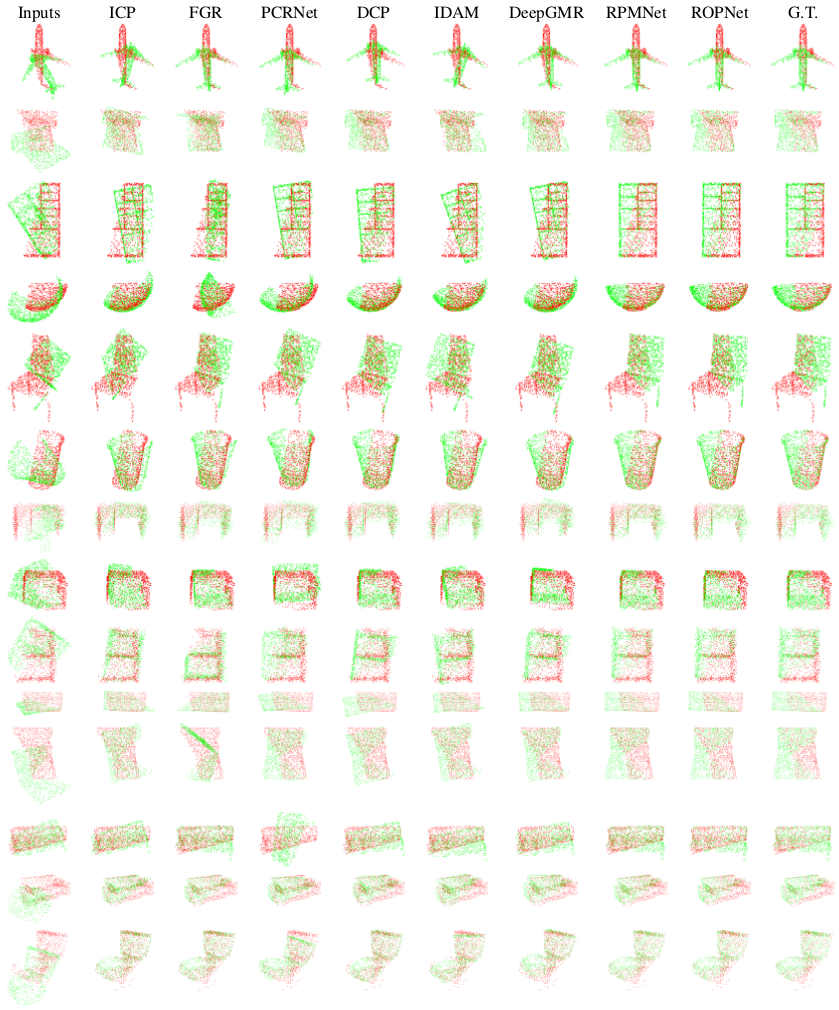}
\caption{Registration results on ModelNet40 unseen shapes. The source and target point clouds $X, Y$ are colored in green and red, respectively}
\label{fig:supp1}
\end{figure}

\paragraph{Unseen shapes}
%FGF: rephrase
%We visualize 14-classes on the ModelNet40 unseen shapes as shown in \autoref{fig:supp1}. 
\autoref{fig:supp1} shows 14-classes of the ModelNet40 unseen shapes.
We select the 10-classes in the first 20 classes in stride two and beginning 0. Also, we select four classes in the last 20 classes. The classes include airplane, bed, bookshelf, bowl, chair, cup, desk, dresser, glass box, keyboard, laptop, piano, sofa, and toilet.

\begin{figure}
\centering
\includegraphics[width=0.98\linewidth]{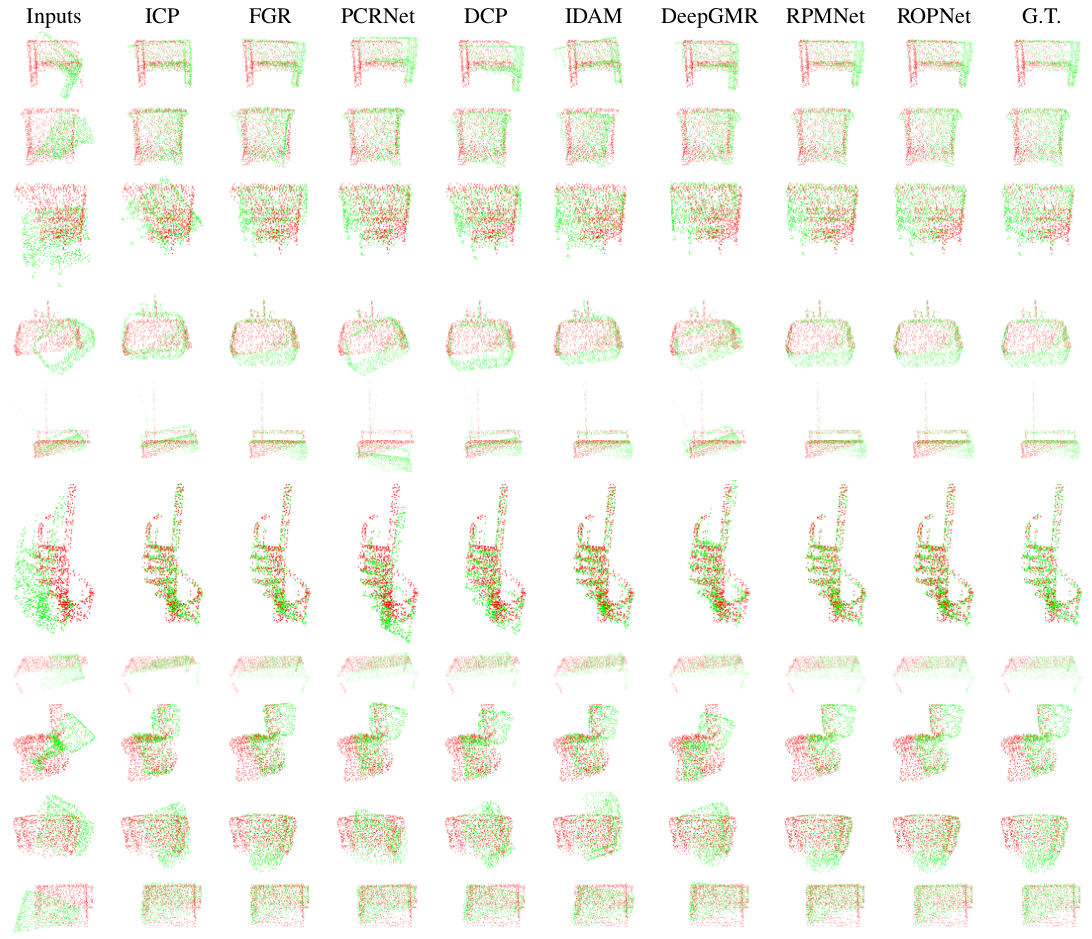}
\caption{Registration results on ModelNet40 unseen categories. The source and target point clouds $X, Y$ are colored in green and red, respectively}
\label{fig:supp2}
\end{figure}

\paragraph{Unseen categories} 
%FGF: rephrase
%We visualize 10-classes on the ModelNet40 unseen categories as shown in \autoref{fig:supp2}. 
\autoref{fig:supp2} shows 10-classes of the ModelNet40 unseen categories.
We select the 10-classes in the last 20 classes in stride two and beginning 21. The classes include mantel, nightstand, piano, radio, sink, stairs, table, toilet, vase, and xbox.

\begin{figure}
\centering
\includegraphics[width=0.98\linewidth]{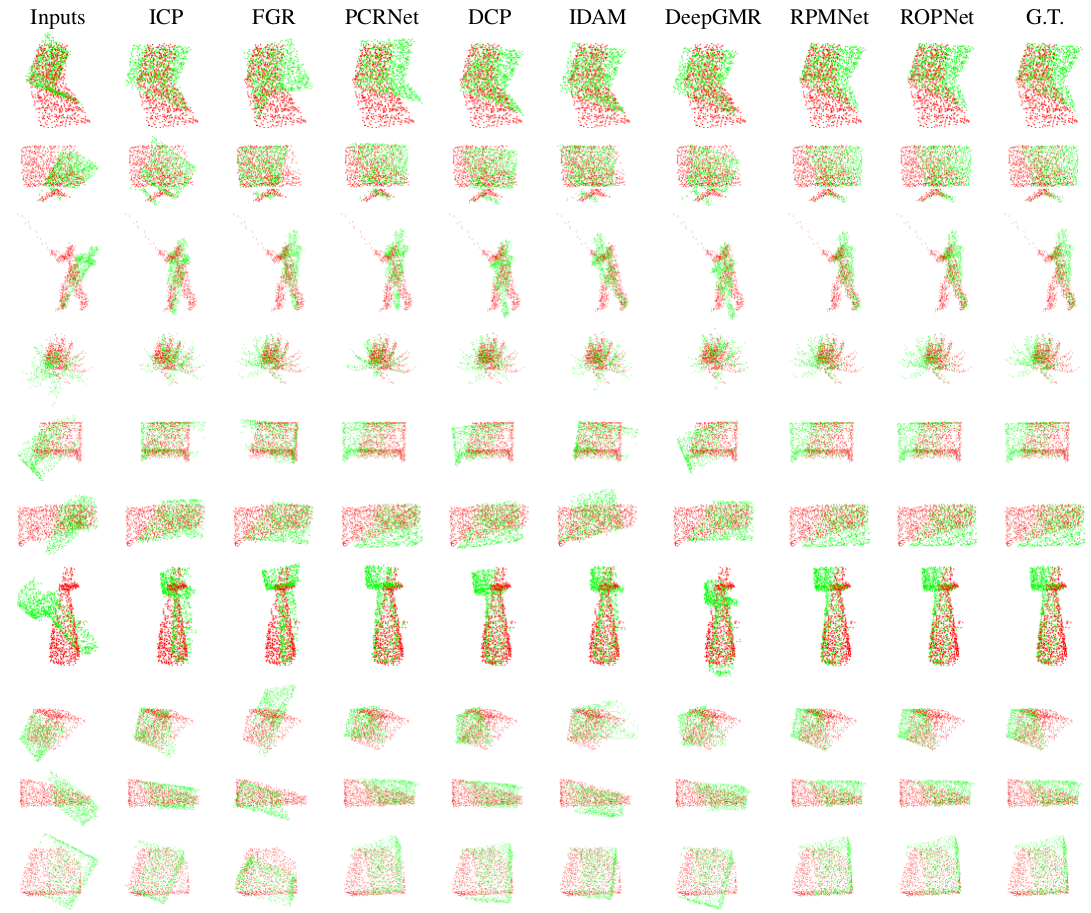}
\caption{Registration results on ModelNet40 unseen categories with Gaussian noise. The source and target point clouds $X, Y$ are colored in green and red, respectively}
\label{fig:supp3}
\end{figure}

\paragraph{Unseen categories with Gaussian noise} 
%FGF: rephrase
%We visualize 10-classes on the ModelNet40 unseen categories with Gaussian noise as shown in \autoref{fig:supp3}. 
\autoref{fig:supp3} shows 10-classes of the ModelNet40 unseen categories with Gaussian noise.
We select the 10-classes in the last 20 classes in stride two and beginning 20. The classes include laptop, monitor, person, plant, range hood, sofa, stool, tent, tv stand, and wardrobe.

\end{document}